\documentclass{article}

\usepackage[utf8]{inputenc} %
\usepackage[T1]{fontenc}    %
\usepackage{
    hyperref,       %
    url,            %
    booktabs,       %
    amsfonts,       %
    nicefrac,       %
    microtype,      %
    mathtools,      %
    silence, 
    }
    
\usepackage{
    xspace,     %
    graphicx,   %
    lipsum,     %
    xargs,      %
    etoolbox,   %
    multirow,
    longtable,
    pdflscape,
    dsfont,   %
    lmodern, %
    amsmath,
    amssymb,
    array,
    placeins, %
    float, %
    xfrac, %
    subcaption,
    makecell, %
}

\usepackage[dvipsnames]{xcolor}
\usepackage[acronym,shortcuts]{glossaries}
\usepackage[inline]{enumitem}
\usepackage[english]{babel}

\usepackage[capitalise, noabbrev]{cleveref} %
\usepackage[export]{adjustbox}

\usepackage{tikz}

\usepackage{subfiles} %

\newcommand{\verylastimports}{%
    \usepackage{csquotes} %
}

\title{The Role of Feature Interactions in Graph-based Tabular Deep Learning}

\author{%
  \name Elias Dubbeldam \email e.f.dubbeldam@uva.nl
  \\ \addr University of Amsterdam
  \AND
  \name Reza Mohammadi \email a.mohammadi@uva.nl
  \\ \addr University of Amsterdam
  \AND
  \name Marit Schoonhoven \email m.schoonhoven@uva.nl
  \\ \addr University of Amsterdam
  \AND
  \name Ilker Birbil \email s.i.birbil@uva.nl
  \\ \addr University of Amsterdam
}

\newcommand{\RR}{\ensuremath{\mathbb{R}}\xspace}
\newcommand{\RRone}{\ensuremath{\mathbb{R}_{[0,1]}}\xspace}

\newcommand{\RRpp}{\in \RR^{p \times p}}
\newcommand{\adj}{adjacency matrix\xspace}

\newcommand{\ft}{FT-Transformer\xspace}
\newcommand{\tg}{T2G-Former\xspace}
\newcommand{\ince}{INCE\xspace}
\newcommand{\fignn}{FiGNN\xspace}
\newcommand{\bdgraph}{BDgraph\xspace}

\newcommand{\nsamples}{N}

\newcommand{\hparam}{hyperparameter\xspace}
\newcommand{\hparams}{hyperparameters\xspace}

\newcommand{\ntrain}{\ensuremath{n_\text{train}}\xspace}
\newcommand{\nvales}{\ensuremath{n_\text{val, early stop}}\xspace}
\newcommand{\nvalhparam}{\ensuremath{n_\text{val, hparam}}\xspace}
\newcommand{\ntest}{\ensuremath{n_\text{test}}\xspace}

\newcommand{\Dtrain}{\ensuremath{D_\text{train}}\xspace}
\newcommand{\Dvales}{\ensuremath{D_\text{val, early stop}}\xspace}
\newcommand{\Dvalhparam}{\ensuremath{D_\text{val, hparam}}\xspace}
\newcommand{\Dtest}{\ensuremath{D_\text{test}}\xspace}

\newcommand{\rscore}{R2 score\xspace}

\newcommand{\cls}{\textsc{cls}\xspace}

\newcommand{\uniform}{\ensuremath{\text{Uniform}}}
\newcommand{\uniformint}{\ensuremath{\text{UniformInt}}}
\newcommand{\loguniform}{\ensuremath{\text{LogUniform}}}
\newcommand{\tpn}[1]{\ensuremath{10^{#1}}}

\newacronym{ml}{ML}{machine learning}
\newacronym{cnn}{CNN}{convolutional neural network}
\newacronym{mse}{MSE}{mean squared error}

\newacronym{tdl}{TDL}{tabular deep learning}
\newacronym{gtdl}{GTDL}{graph-based tabular deep learning}
\newacronym{rdl}{RDL}{relational deep learning}

\newacronym{gnn}{GNN}{graph neural network}
\newacronym{gbdt}{GBDT}{gradient boosted deep trees}
\newacronym{tpe}{TPE}{tree-structured Parzen estimator}
\newacronym{pgm}{PGM}{probabilistic graphical model}

\newacronym{frgraph}{FR-Graph}{feature-relation graph}

\newacronym{roc}{ROC AUC}{receiver operating characteristic area under curve}

\newacronym{scm}{SCM}{structural causal model}
\newacronym{dag}{DAG}{directed acyclic graph}
\newacronym{mvn}{MVN}{multivariate normal}

\newacronym{sota}{SOTA}{state of the art}

\newcommand{\gtdl}{\gls{gtdl}\xspace}

\newcommand{\gnn}{\gls{gnn}\xspace}
\newcommand{\gnns}{\glspl{gnn}\xspace}
\newcommand{\pgm}{\gls{pgm}\xspace}
\newcommand{\pgms}{\glspl{pgm}\xspace}

\newcommand{\frgraph}{\gls{frgraph}\xspace}

\newcommand{\scm}{\gls{scm}\xspace}
\newcommand{\mvn}{\gls{mvn}\xspace}
\newcommand{\roc}{\gls{roc}\xspace}
\newcommand{\mse}{\gls{mse}\xspace}

\def\month{02}
\def\year{2026}
\def\openreview{\url{https://openreview.net/forum?id=olGaiwoZHZ}}

\usepackage[accepted]{utils/style/tmlr/tmlr} %

\providecommand{\subfix}[1]{#1}

\def\insertbib{\ifSubfilesClassLoaded{%
    \bibliography{\subfix{../files/PhD_BibTeX}}%
    \bibliographystyle{utils/style/tmlr/tmlr}%
}{}}

\def\insertbibmain{%
    \bibliography{files/PhD_BibTeX}%
    \bibliographystyle{utils/style/tmlr/tmlr}%
}

\graphicspath{{figures/}}

\newcommandx{\eliascaption}[2][1=]{\todo[inline,linecolor=blue,backgroundcolor=blue!25,bordercolor=blue,#1]{\textit{Elias:} #2}}

\newcommandx{\ilkercaption}[2][1=]{\todo[inline,linecolor=Green,backgroundcolor=Green!25,bordercolor=Green,#1]{\textit{Ilker:} #2}}

\newcommandx{\maritcaption}[2][1=]{\todo[inline,linecolor=red,backgroundcolor=red!25,bordercolor=red,#1]{\textit{Marit:} #2}}

\newcommandx{\rezacaption}[2][1=]{\todo[inline,linecolor=Fuchsia,backgroundcolor=Fuchsia!25,bordercolor=Fuchsia,#1]{\textit{Reza:} #2}}

\setlipsum{%
  par-before = \begingroup\color{gray},
  par-after = \endgroup
}
\glsdisablehyper
\crefname{subsection}{Subsection}{Subsections}

\newbool{hide_custom_warnings}
\boolfalse{hide_custom_warnings}
\AtBeginDocument{%
  \ifbool{hide_custom_warnings}{%
    \WarningFilter{latex}{Reference}
    \WarningFilter{latex}{Marginpar on page}
    \WarningFilter{latex}{Float specifier changed}
    \WarningFilter{latex}{Float specifier changed} %
    \hbadness=9999999
    \WarningFilter{chktex}{Command terminated with space.}
    \WarningFilter{latex}{You have requested package}
  }{%
  }%
}

\newbool{show_outline}
\boolfalse{show_outline} 
\newcommand{\outline}[1]{%
  \ifbool{show_outline}%
    {\textcolor{purple}{[#1] }}%
    {}%
}

\newbool{fitemize}
\booltrue{fitemize}
\newenvironment{fitemize}
  {%
    \ifbool{fitemize}
      {%
        \begin{itemize*}[afterlabel={}, label={}, itemjoin={\space}]%
      }
      {%
        \begin{itemize}[noitemsep, nolistsep]%
      }%
  }
  {%
    \ifbool{fitemize}
      {\end{itemize*}}
      {\end{itemize}}%
  }

\verylastimports %

\begin{document}

\maketitle

\begin{abstract}

Accurate predictions on tabular data rely on capturing complex, dataset-specific feature interactions.
Attention-based methods and graph neural networks, referred to as graph-based tabular deep learning (GTDL), aim to improve predictions by modeling these interactions as a graph.
In this work, we analyze how these methods model the feature interactions.
Current GTDL approaches primarily focus on optimizing predictive accuracy, often neglecting the accurate modeling of the underlying graph structure.
Using synthetic datasets with known ground-truth graph structures, we find that current GTDL methods fail to recover meaningful feature interactions, as their edge recovery is close to random.
This suggests that the attention mechanism and message-passing schemes used in GTDL do not effectively capture feature interactions.
Furthermore, when we impose the true interaction structure, we find that the predictive accuracy improves.
This highlights the need for GTDL methods to prioritize accurate modeling of the graph structure, as it leads to better predictions.

\end{abstract}

\section{Introduction}
Deep learning has achieved remarkable success in domains such as natural language processing and computer vision. On tabular data, however, deep learning methods still struggle to compete against traditional, tree-based machine learning methods \citep{grinsztajnWhyTreebasedModels2022, mcelfreshWhenNeuralNets2024}.
Although recent advances in tabular deep learning occasionally surpass these baselines on select benchmarks (e.g., \citet{gorishniyRevisitingDeepLearning2021, hollmannAccuratePredictionsSmall2025}), no deep learning method has yet demonstrated consistent superiority across datasets and evaluation settings \citep{grinsztajnWhyTreebasedModels2022, mcelfreshWhenNeuralNets2024}.

\outline{Why feature interactions?} Tabular data is characterized by the heterogeneous nature of its features: each feature often encodes distinct semantics, and relationships among features (or feature interactions) can be complex, indirect, and dataset-specific. 
By modeling the feature interactions, one incorporates the \textit{inductive bias} (domain-specific principles embedded into the model's architecture \citep{goyalInductiveBiasesDeep2022, battagliaRelationalInductiveBiases2018, romeroguzmanGoodEfficientInductive2024}) that features interact with each other differently. With `modeling' we mean that the network has, by design, separate parameters for each feature interaction.
Using inductive biases has proven to be important for success in other fields of deep learning.
For example, \glspl{cnn} achieve sample efficiency and robustness in computer vision by encoding translational invariance \citep{lecunBackpropagationAppliedHandwritten1989, fukushimaNeocognitronSelforganizingNeural1980}, and transformers excel in natural language processing using attention-based mechanisms to capture sequential and contextual relationships \citep{vaswaniAttentionAllYou2017}.

\outline{Graphs} Modeling this inductive bias of feature interactions comes naturally in the form of a graph, where the nodes represent features and the edges their interactions. 
\Glspl{pgm} have a rich history in statistics, providing a framework to model multivariate dependencies \citep{lauritzenGraphicalModels1996}. 
These methods excel at robustly describing the graph structure while enabling predictions, yet they lack the ability to model complex nonlinear relationships that deep learning can provide.
\Gls{gtdl} methods aim to merge the expressive power of deep learning with graph-structured feature representations.
\textit{Feature} \gnns, reviewed by \citet{liGraphNeuralNetworks2024}, are \gnns focused on tabular data, having the features as nodes and the feature interactions as edges.\footnote{This categorization of feature graphs contrasts with instance graphs, where nodes represent instances (rows) and edges capture relationships between those instances.}
However, how these feature interactions are modeled within these networks has not been extensively studied or evaluated. These methods do not evaluate explicitly whether their learned feature interactions accurately correspond to meaningful relationships in the data.

\begin{figure}[htbp]
  \centering
  \includegraphics[width = 0.9 \linewidth]{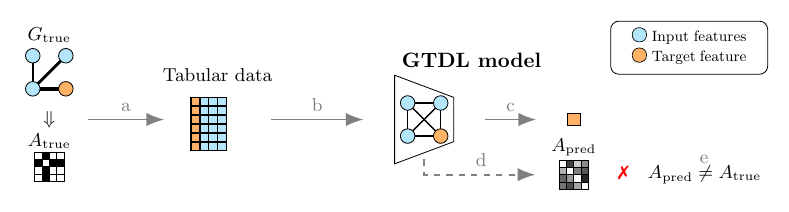}
  \caption{
    (a) A true underlying graph structure generates tabular data.
    (b, c) Only the data is used to train a \gtdl model (using a fully connected graph) to predict the target feature. 
    (d) After training, the learned graph structure is extracted from the model. 
    (e) When this predicted graph structure is compared to the true graph structure, we find that they are not similar for existing \gtdl methods.
    }
    \label{fig:problem_statement}
  \end{figure}

Existing \gtdl methods (e.g., \citet{
  liFiGNNModelingFeature2020,
  yanT2GFormerOrganizingTabular2023,
  zhengDeepTabularData2023,
  villaizan-valleladoGraphNeuralNetwork2024,
  yeCrossFeatureInteractiveTabular2024,
zhouTable2GraphTransformingTabular2022}) typically evaluate the learned graph structure only qualitatively, as real-world datasets rarely include ground-truth feature interaction graphs.
Their training loss is tied to predictive performance, providing no incentive to ensure accuracy or meaningfulness in the learned graph structure. As a result, the adjacency matrix may reflect optimization artifacts rather than genuine feature interactions, as sketched in \cref{fig:problem_statement}.
This emphasis on predictive metrics over structural fidelity constrains interpretability. Overall, there remains a lack of systematic techniques for both validating learned feature interactions and guiding learning with prior knowledge.

In this work, we analyze whether existing \gtdl methods learn meaningful feature interactions. 
Do \gtdl methods learn an accurate graph structure when being trained to predict a target feature? 
Conversely, does an exclusive focus on predictive accuracy lead to spurious interactions rather than genuine feature dependencies, thereby undermining robustness, generalization, and explainability? 
When the graph structure is modeled correctly, does the predictive performance of \gtdl methods improve?

\outline{Outline} To support our analysis, we take the following steps.
In \cref{sec:related-work}, we review the existing literature on \gtdl methods and identify their limitations in the evaluation and validation of learned feature interactions. 
In \cref{sec:method}, we introduce a framework to systematically evaluate how well predictive \gtdl methods learn the graph structure. The two key parts of this framework are (i) synthetic datasets with known ground-truth graphs, and (ii) a metric to quantitatively assess the accuracy of learned graphs.
In \cref{sec:analysis}, we discuss the results of controlled experiments to showcase the framework. Specifically, we show that existing \gtdl methods fail to recover meaningful feature interactions, and that enforcing the true interaction structure improves predictive performance.
In \cref{sec:conclusion}, we give a concluding discussion and propose future research directions.

\insertbib

\section{GTDL methods and their feature interactions} \label{sec:related-work}

After formalizing the problem context, we organize this section around methods that model feature interactions in tabular data through feature graphs. 
First, we review attention-based methods and \gnns for tabular data, and discuss how both are interpreted as \gtdl methods.
Next, we relate \gtdl to \pgms, which serve as a principled baseline for learning conditional independence structure.
Lastly, we investigate methods that may appear related at first glance but, upon closer examination, are not directly pertinent to this study.

\paragraph{Problem setting and notation.}
A full tabular dataset $D = \left[x \parallel y\right] \in \RR^{n \times p}$ consists in a traditional supervised setting of input features $x \in \RR^{n \times (p-1)}$ and a target feature $y \in \RR^{n \times 1}$, with $n$ the number of samples, $p$ the number of features and $\parallel$ indicating concatenation. When we refer to `features', we mean both input and target features.
Features could be either numerical or categorical.

The features and their interactions can be represented as an undirected graph $G = (V, E)$, where $V$ is the set of nodes and $E$ the set of edges, such that the number of nodes $|V| = p$.
A binary symmetric adjacency matrix $A_\text{true}$ describes the true graph structure. The absence of an edge between two nodes indicates the conditional independence of these two nodes conditioned on all other nodes (this is different from directed causal graphs, where the presence of an edge indicates a direct causal effect between two nodes.)

From trained \gtdl methods, we can extract a weighted adjacency matrix $A \in \RR^{p \times p}$, where $0 \leq A_{ij} \leq 1$ indicates the strength of the interaction between features $i$ and $j$, with $i,j\in \{1, \ldots, p\}$.
As the interaction from feature $i$ to $j$ could be different from the interaction from feature $j$ to $i$, the weighted adjacency matrix is not necessarily symmetric:
While \(A_{ij}\) can differ from \(A_{ji}\), they should be symmetric in support (i.e., they share the same sparsity pattern). 
Specifically, if \(A_{\text{true},ij} = 0\), then it should be that \(A_{ij}, A_{ji} = 0\). 
Conversely, if \(A_{\text{true},ij} = 1\), then it should be that \(A_{ij}, A_{ji} > 0\).

The main task is to predict the target feature \(y\) given the input features \(x\). Correct modeling of the feature interactions could improve this prediction, including both target-input interactions (relationships between \(y\) and each \(x_i\)) and input-input interactions (relationships between pairs of input features \(x_i\) and \(x_j\)). The strength of feature interactions between two nodes should be (close to) zero when these nodes are conditionally independent.

\subsection{Attention-based methods} \label{sec:attention-based}

Due to the success of the transformer architecture \citep{vaswaniAttentionAllYou2017}, most recent tabular deep learning methods are attention-based \citep{arikTabNetAttentiveInterpretable2020, huangTabTransformerTabularData2020, somepalliSAINTImprovedNeural2021, kossenSelfAttentionDatapointsGoing2021, gorishniyRevisitingDeepLearning2021}, from which \ft \citep{gorishniyRevisitingDeepLearning2021} has been established as a popular baseline. 
All of these methods are based on multi-head self-attention \citep{vaswaniAttentionAllYou2017}.

In most works, the attention map is of size $p \times p$ if a target token (or \cls, `classification' or `output' token \citep{devlinBERTPretrainingDeep2019}) is appended. %
If the attention map is equal to the size of the total number of features (or the number of input features), it can be used for interpretation and explaining the feature interactions. 
This approach is popular in natural language, with BertViz \citep{vigMultiscaleVisualizationAttention2019} being used to interpret how the model assigns weights to different tokens.
While in natural language the size of attention map differs per input due to varying sequence lengths, in tabular data the number of features is fixed, making the attention map the same size as the adjacency matrix. This allows for an interpretation of the attention map as a proxy for the weighted adjacency matrix, which is further discussed in \cref{sec:eval}.

Therefore, we refer to such attention-based methods as \textit{implicit} \gtdl methods. They do not explicitly model the feature interactions, but due to the nature of the attention map, they do model the graph structure implicitly.

This notion, that the attention map can be interpreted as the learned graph structure of tabular data, has not been thoroughly discussed in the literature. However, most methods, as listed in \cref{tab:literature_overview}, use the attention map for interpretability.
\begin{fitemize}
  \item TabNet \citep{arikTabNetAttentiveInterpretable2020} reports the attention map of synthetic datasets as visualizations and notes how irrelevant features are ignored in the attention map. 
  \item SAINT \citep{somepalliSAINTImprovedNeural2021}, although designed for tabular data, reports the attention map of MNIST and discusses that the visualization of the attention map is similar to the ground-truth image.
  \item \ft \citep{gorishniyRevisitingDeepLearning2021} interprets the attention map as feature importances, and shows for some real-world datasets that the attention map has a high rank correlation with integrated gradients \citep{sundararajanAxiomaticAttributionDeep2017}, a method to measure feature importance.
\end{fitemize}

\begin{table}[htbp]
\caption{
\Gls{gtdl} methods evaluate the feature interaction only qualitatively, typically with a visualization of the attention map or adjacency matrix.
Size denotes the length of square attention map or adjacency matrix $A$ (e.g., size $p$ means $A \in \RR^{p \times p}$).
}
\label{tab:literature_overview}
\centering
\begin{tabular}{llll}
  \toprule
  \textbf{Model} & \textbf{Reference} & \textbf{Size} & \textbf{Feature interaction evaluation} \\ %
  \midrule
  \multicolumn{3}{l}{\textbf{Attention-based}} \\
  \midrule
  FT-Transformer & \citet{gorishniyRevisitingDeepLearning2021} & $p$ & Correlation with feature importance \\ %
  TabNet & \citet{arikTabNetAttentiveInterpretable2020} & $p$ & Visual of synthetic dataset \\ %
  SAINT & \citet{somepalliSAINTImprovedNeural2021} & $p$ & Visual of MNIST \\ %

  \addlinespace[0.5ex]
  \midrule
  \multicolumn{3}{l}{\textbf{Graph neural network}} \\
  \midrule
  \fignn & \citet{liFiGNNModelingFeature2020} & $p-1$  & Visual of real-world dataset \\ %
  \tg & \citet{yanT2GFormerOrganizingTabular2023} & $p$ & Visual of real-world datasets \\ %
  DRSA-Net & \citet{zhengDeepTabularData2023} & $p-1$ & Visual of real-world dataset \\ %
  \ince & \citet{villaizan-valleladoGraphNeuralNetwork2024} & $p$ & Visual of real-world dataset \\ %
  MPCFIN & \citet{yeCrossFeatureInteractiveTabular2024} & $p$ & Visual of real-world datasets \\ %
  Table2Graph & \citet{zhouTable2GraphTransformingTabular2022} & $p-1$ & Visual synthetic dataset \\ %
  \bottomrule
\end{tabular}
\end{table}

\subsection{\Glsentrylongpl{gnn}} \label{sec:gnn}
\Glspl{gnn} operate directly on graph-structured data by propagating information between connected nodes \citep{zhouGraphNeuralNetworks2021}. 
Feature \gnns apply this paradigm to tabular data, modeling each feature as a node and explicitly learning feature interactions through message passing \citep{liGraphNeuralNetworks2024}.

Generally, the message-passing mechanism can be summarized as
\begin{equation} \label{eq:gnn_mp}
  h_i^{(l)} = \phi^{(l)} \left( h_i^{(l-1)}, \bigoplus_{j \in \mathcal{N}(i)} \psi^{(l)} \left( h_i^{(l-1)}, h_j^{(l-1)}, A_{ij} \right) \right),
\end{equation}
with 
$h_i^{(l)}$ the representation of node $i$ at layer $l$, 
$\mathcal{N}(i)$ the neighbors of node $i$, \footnote{In \gtdl, the neighbors are typically all other features, i.e., $\mathcal{N}(i) = \{1, \ldots, p\} \setminus \{i\}$. A fully connected graph is used due to the absence of a known graph structure.}
$\psi^{(l)}$ a message function, $\bigoplus$ an aggregation operator, and $\phi^{(l)}$ an update function. 
The details of these differ between \gnn architectures.
Attention-based methods can be seen as a special case of \gnns \citep{joshiTransformersAreGraph2025}.
In the context of tabular data, a key advantage of \gnns is that they generalize attention-based methods by using a trainable weighted adjacency matrix, $A$, to explicitly propagate information between nodes—rather than relying on implicitly learned attention maps. Furthermore, \gnns for tabular data typically have trainable parameters that represent individual features or interactions, allowing for more flexible modeling of feature interactions compared to attention-based methods that rely on shared parameters across features.

Therefore, we refer to them as \textit{explicit} \gtdl methods, contrary to attention-based methods that model the graph structure implicitly.

Different \gtdl methods implement \cref{eq:gnn_mp} in different ways.
\begin{fitemize}
  \item \fignn \citep{liFiGNNModelingFeature2020} uses a feature graph to explicitly model the separate feature interactions.
  \item \tg \citep{yanT2GFormerOrganizingTabular2023} adapts the transformer architecture \citep{vaswaniAttentionAllYou2017} for tabular data and learns a feature graph that focuses on learning meaningful interaction between different features. 
  \item DRSA-Net \citep{zhengDeepTabularData2023} uses dual-route structure \gnns to learn adaptively the sparse graph structure.
  \item \ince \citep{villaizan-valleladoGraphNeuralNetwork2024} has a similar approach as \tg, but uses an Interaction Network \citep{battagliaInteractionNetworksLearning2016} instead of a Transformer. 
  \item MPCFIN \citep{yeCrossFeatureInteractiveTabular2024} uses cross-feature embeddings and multiplex \gnns to model the interaction and dependencies between features.
\end{fitemize}

The feature \gnn literature \citep{
  liFiGNNModelingFeature2020,
  yanT2GFormerOrganizingTabular2023,
  zhengDeepTabularData2023,
  villaizan-valleladoGraphNeuralNetwork2024,
  yeCrossFeatureInteractiveTabular2024,
zhouTable2GraphTransformingTabular2022} suggests that the learned adjacency matrix can be used to interpret and explain the feature interactions. However, there are two reasons to be careful with this interpretation.
\begin{enumerate}[noitemsep, topsep=0pt]
  \item 
  The evaluation of the graph structure is only heuristic to the best of our knowledge.
  The aforementioned explicit \gtdl methods (\fignn, \ince, \tg, MPCFIN, and DRSA-Net) report the learned adjacency matrix for one or a few real-world datasets. They argue that the learned feature interactions are meaningful by post-hoc explaining the feature interactions. 
  They justify the connections by referencing the semantic meaning of the feature names, suggesting that connected features are intuitively related.
  The issue with this approach is that, in the absence of a ground-truth graph structure, it becomes impossible to quantitatively evaluate the learned adjacency matrix.

  \item
  The GTDL methods do not explicitly instruct the model to learn the true underlying graph structure. 
  The loss is computed exclusively on the error between predicted and true target values. This means the model is only incentivized to improve predictive accuracy, not to accurately model the underlying feature interactions. As a result, the learned graph structure may not reliably reflect the true relationships between features. This limits its usefulness for interpretability and potentially constrains predictive performance.
\end{enumerate}

Table2Graph \citep{zhouTable2GraphTransformingTabular2022} addresses the first problem, that real-world datasets do not have a ground truth graph structure, by using a synthetic dataset.
However, the learned graph structure is still evaluated heuristically against the ground-truth interactions, by visually comparing the learned weighted adjacency matrix with the ground-truth interactions.
The second problem, that the model is only prediction-centric, is addressed by introducing a reinforcement learning term to the loss function to explore the adjacency matrix.
This encourages the model to also focus on learning the graph structure, rather than just the predictive performance.

\outline{Node-level and graph-level task} 
Not all \gnn methods treat predicting the target feature in the same way. Some of these models treat predicting the target feature as a node-level task, while others treat it as a graph-level task \citep{prince2023understanding}. Node-level approaches (\tg, \ince, MPCFIN) include a target node in the graph structure, and pass the embedding of the target node to an output layer. With this approach, the model learns a weighted adjacency matrix of size $p \times p$. Graph-level approaches (\fignn, DSRA-Net, Table2Graph) do not include a target node, resulting in an \adj of size $(p-1) \times (p-1)$. The embeddings of all nodes are aggregated and passed to an output layer.

\subsection{Probabilistic graphical model as a baseline for GTDL} \label{sec:pgm}

Feature graphs, as studied by \gtdl methods, share similarities with \pgms \citep{lauritzenGraphicalModels1996, kollerProbabilisticGraphicalModels2009}. \Glspl{pgm} provide a principled framework for modeling multivariate dependencies by encoding conditional independence relationships among random variables using graphs. Widely applied in Bayesian statistics, \pgms represent the structure of a probability distribution, often Gaussian, through a compact graph encoding the conditional independencies among variables. 
Bayesian techniques in \pgms, like \bdgraph \citep{mohammadiBDgraphPackageBayesian2019}, have demonstrated strong empirical performance in recovering interaction structures \citep{vogelsBayesianStructureLearning2024}. 
The ability to quantify uncertainty in the learned graph structure makes them a useful sanity check for GTDL methods.

\subsection{Related approaches} \label{sec:outside_scope}

There are other related approaches that are after closer inspection not relevant to our discussion on \gtdl.
In the literature on recommender systems and click-through rate,  there has been a longer interest in a different notion of feature interactions, that of \textit{cross features}; e.g., \citep{chengWideDeepLearning2016, guoDeepFMFactorizationMachineBased2017, lianXDeepFMCombiningExplicit2018, wangDeepCrossNetwork2017, caiARMNetAdaptiveRelation2021, wangDCNV2Improved2020, songAutoIntAutomaticFeature2019}. These models focus on learning multiple weighted products of features to improve the prediction of the target feature.
As noted by \citet{liFiGNNModelingFeature2020}, this limits the capability to model interactions across different features flexibly and explicitly. We are interested in how to model the feature interactions explicitly on a graph. Therefore, we do not discuss the feature interactions of these methods in further detail.

Tabular foundation models, \begin{fitemize}
  \item TabPFN \citep{hollmannAccuratePredictionsSmall2025},
  \item TabICL \citep{quTabICLTabularFoundation2025} and
  \item LimiX \citep{zhangLimiXUnleashingStructuredData2025}
\end{fitemize}
have recently gained attention due to their high predictive performance on tabular data. Two key aspects of these models are their alternating instance-wise and feature-wise attention layers, and their ensembling predictions over multiple feature permutations.
The feature-wise attention layers work similar as the attention-based methods discussed in \cref{sec:attention-based}, and therefore could have been interpreted as implicit \gtdl methods. However, these tabular foundation models contain techniques that prevent a straightforward interpretation of the feature-wise attention map as a weighted adjacency matrix. TabPFN and LimiX encode groups of features collectively rather than individually. TabICL incorporates rotary positional embedding (RoPE) \citep{suRoFormerEnhancedTransformer2024} independent of the feature permutation, which alters the attention map. Therefore, we do not consider the feature interactions of these models in this work.

Tree-based models (e.g., \begin{fitemize}
  \item XGBoost \citep{chenXGBoostScalableTree2016}, 
  \item LightGBM \citep{keLightGBMHighlyEfficient2017} and
  \item CatBoost \citep{prokhorenkovaCatBoostUnbiasedBoosting2019}
\end{fitemize}) remain a popular choice for tabular data. 
Nevertheless, these approaches do not explicitly represent feature interactions in a graphical format.
Due to the nature of tree architectures, the learned feature interactions can not easily be extracted from the model.

\insertbib

\section{Evaluating feature interactions in GTDL}\label{sec:method}

The development of \gtdl is hindered by the fact that the learned graph structure is only evaluated heuristically.
To solve this, we introduce a framework to evaluate the learned graph structure of \gtdl methods with synthetic datasets and quantitative metrics.

\subsection{Synthetic data with a graph structure} \label{sec:dataset}

\outline{Problem with real-world datasets} 
Most existing \gtdl methods lack rigorous evaluation of the learned graph structure. Typically, the learned graph structure is evaluated heuristically, by reporting a visualization of the learned weighted \adj of real-world datasets. Feature interactions are post-hoc explained based on the semantic meaning of the feature names. 
Evaluating only on real-world datasets is problematic, as the \textit{true} graph structure is not known.
Therefore, we propose using \textit{synthetic} datasets.
Using synthetic data enables \gtdl methods to compare the learned graph structure with the ground-truth underlying graph structure in a controlled environment.

\begin{figure}[htbp]
    \centering
    {\small %
    \begin{tabular}{
        l @{\hspace{0.2em}} 
        l @{\hspace{0.4em}} 
        c @{\hspace{0.2em}}
        c @{\hspace{0.2em}}
        l 
        |
        l @{\hspace{0.2em}}
        l @{\hspace{0.4em}}
        c @{\hspace{0.2em}}
        c @{\hspace{0.2em}}
        l @{\hspace{0.2em}}
        }
        \multicolumn{5}{c|}{\textbf{Multivariate normal}} & \multicolumn{5}{c}{\textbf{Structural causal model}} \\
        
        {\scriptsize(i)} & $G_\text{true}$           & $=$    & \includegraphics[height=3.em,valign=c]{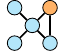}             & $\sim \text{Bernoulli}$ &
        {\scriptsize(i)} & DAG & $=$ & \includegraphics[height=3.em,valign=c]{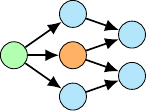} &  $\sim \text{DAG generator}$  \\[2.5ex]
        & & & & & 
        & $G_\text{true}$       & $=$    & \includegraphics[height=3.em,valign=c]{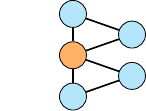} & $= $ \makecell[l]{moralized and \\ marginalized DAG} \\
        
        {\scriptsize(ii)} & $\Sigma_G$      & $=$    & \includegraphics[height=2.5em,valign=c]{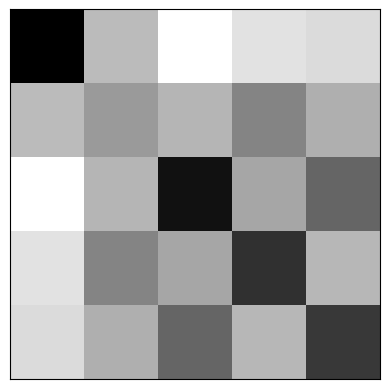} & $\sim \text{G-Wishart}$ & 
        {\scriptsize(ii)} & $f_i(x_{i, \text{parents}})$     & $=$ & \includegraphics[height=1.5em,valign=c]{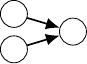}   & $\sim$ \makecell[l]{computational map \\ generator} \\[2.5ex]
        {\scriptsize(iii)} & $D$           & $\in$    & $R^{n \times p}$ & $\sim \mathcal{N}(0, \Sigma_G)$ & 
        {\scriptsize(iii)} & $x_\mathrm{roots}$           &   &  & $\sim \mathcal{N}(0, \sigma)$ \\
        & & & & & 
        & $D$           & $\in$    & $R^{n \times p}$ & $= \mathrm{SCM}_{\text{DAG}, \{f_i\}}(x_\mathrm{roots})$
    \end{tabular}
    }
    \caption{Two synthetic data generation pipelines.
    Both pipelines can roughly be divided into three steps.
    (i) Sample a graph structure.
    (ii) Sample feature interaction.
    (iii) Sample data given the graph and feature interactions.
    Nodes are colored as {\color{cyan!30}\textbullet} cyan input features $x$, {\color{orange!60}\textbullet} orange target feature $y$, and {\color{green!60}\textbullet} green root nodes $x_\mathrm{roots}$. 
    }
    \label{fig:synthetic_datagen}
\end{figure}

We adapt two existing data generation methods from the literature. The three-step process is sketched in \cref{fig:synthetic_datagen}, and the details are given in \cref{sec:data_gen}.
\begin{itemize}[topsep=0pt, leftmargin=*]
    
    \item \textbf{\Glspl{mvn}}
    are typically studied by \pgm methods. We follow the default procedure of generating conditional multivariate data (as described in \citep{mohammadiBayesianStructureLearning2015}, for instance). In short, we employ the following:
    \begin{enumerate*}[label=(\roman*)]
        \item Sample a graph structure $G_\text{true} \RRpp$ from the Bernoulli distribution.
        \item Sample a covariance matrix $\Sigma_G \RRpp$ from the G-Wishart distribution \citep{roveratoHyperInverseWishart2002, letacWishartDistributionsDecomposable2007} to describe the feature interactions.
        \item Obtain $n$ samples $D \in \RR^{n \times p}$ from $\mathcal{N}(0, \Sigma_G)$.
    \end{enumerate*}
    
    \item \textbf{\Gls{scm}} \citep{pearlCausalityModelsReasoning2021} are used to generate tabular data in tabular foundation models \citep{hollmannAccuratePredictionsSmall2025, quTabICLTabularFoundation2025, zhangLimiXUnleashingStructuredData2025}.
    We follow a simplified version of the data generation process in these tabular foundation models, that is:
    \begin{enumerate*}[label=(\roman*)]
        \item Generate a \gls{dag} to define the graph structure of the \scm. To obtain the undirected graph $G_\text{true}$ from the \gls{dag}, we moralize (connect all parents of a child node, and drop the direction of the edges because the child of a parent can also be used to predict the parent) and marginalize (drop the root nodes and connect its children) the \gls{dag} \citep{cowellProbabilisticNetworksExpert1999}. More details on moralization and marginalization are discussed in \cref{sec:data_gen}.
        \item Sample computational maps $f_i$ for each child node $i$ in the \gls{dag}. The computational maps $\{f_i\}$ are smooth nonlinear functions that take all the values of the incoming edges as input, and the output is the value of the child node $i$.
        \item Traverse random data $x_\text{roots}$ in topological order through the \gls{dag} to obtain $n$ samples $D \in \RR^{n \times p}$. 
    \end{enumerate*}
\end{itemize}

\outline{What should the model learn} For both approaches, we randomly select a target feature $y \in \RR^{n \times 1}$ from $D$, with the remaining columns serving as input features $x \in \RR^{n \times (p-1)}$. 
The target feature is directly influenced by its neighbors and only indirectly by non-neighbors. This setup lends itself well for evaluating the graph structure learned by \gtdl methods.

Consider the simple example: $x_0 - x_1 - x_2$. The model could learn to use $x_0$ to predict $x_2$ directly by learning an edge between them. However, this is suboptimal because $x_2$ is conditionally independent of $x_0$ given $x_1$. That is, once $x_1$ is known, $x_0$ provides no additional information for predicting $x_2$ \citep{lauritzenGraphicalModels1996}.
When observations are noisy, using $x_0$ to predict $x_2$ accumulates uncertainty across multiple dependencies, leading to worse predictions than those obtained by using the immediate neighbor $x_1$.
The model should instead recognize $x_1$ as a more reliable predictor for $x_2$, which results in learning the correct graph structure. 

The proposed synthetic data generation methods are different from the synthetic datasets used by TabNet \citep{arikTabNetAttentiveInterpretable2020} or Table2Graph \citep{zhouTable2GraphTransformingTabular2022}, previously discussed in \cref{sec:attention-based,sec:gnn}. All input features in these datasets are conditionally independent of each other, and have a direct interaction with the target feature. Therefore, there is no underlying graph structure to be learned, as all features are connected only to the target feature. In contrast, the \mvn and \scm data generation methods create datasets with more complex underlying graph structures, making them more suitable to evaluate \gtdl methods.

We acknowledge that our data generation processes may not be fully representative of real-world tabular data. For instance, the graphs evaluated in this work (\cref{sec:experiment,sec:analysis}) are relatively small ($p = 10$), the \mvn does only have linear feature interactions, and the \scm does not have missing nodes within the \gls{dag}. However, key is that the datasets have a clear underlying ground-truth graph structure that \gtdl methods should be able to learn. If models can not model the feature interactions of these synthetic datasets well, it is unlikely to expect that, on larger ($p \gtrsim 100$) and real-world datasets, these models will learn meaningful feature interactions.

\subsection{Interpreting and evaluating the graph structure} \label{sec:eval}

The \gtdl methods introduced in \cref{sec:attention-based,sec:gnn} model the graph structure inside their architecture. In this section, we discuss how we extract the learned weighted \adj $A \in \RR^{p \times p}$ from these methods, and how we propose to evaluate the quality of the learned graph structure. For the \gnn-based \gtdl methods, extracting the \adj is straightforward, as these methods explicitly use a weighted \adj within their architecture. For the attention-based \gtdl methods, we interpret the attention maps as a proxy for the learned weighted \adj.

To justify this interpretation, consider the following example: Let $a$ be a $p \times p$ matrix representing the attention map, where each element $a_{ij}$, denotes the attention weight from feature $i$ to feature $j$, as used by \citet{vaswaniAttentionAllYou2017}.
If feature $i$ is conditionally independent of feature $j$ given all other features, we have $A_{\text{true},ij} = 0$. The attention mechanism should learn during training to assign low attention weights $a_{ij}, a_{ji} \approx 0$, as feature $i$ does not rely on information from feature $j$ (and vice versa) for its representation. 
Contrary, if feature $i$ and feature $j$ are dependent, we have $A_{\text{true},ij} = 1$. The attention mechanism should learn to assign non-zero attention weights $a_{ij}, a_{ji} > 0$, as feature $i$ relies on information from feature $j$ (and vice versa) for its representation.
In summary, with $A_{\text{true},ij} = 0$ we expect $a_{ij}, a_{ji} \approx 0$, and with $A_{\text{true},ij} = 1$, we expect $a_{ij}, a_{ji} > 0$.

After training, the attention maps of the test samples are extracted for all heads and layers. To obtain the adjacency matrix we perform two steps. First, we average the attention maps over the test samples, heads and layers to obtain and to obtain a single attention map of size $p \times p$. Second, we account for the diagonal of the attention map and for the softmax normalization from the original attention equation, which is further explained in \cref{sec:lit_implementation}, to obtain the weighted adjacency matrix $A$.

Current \gtdl methods only report the predictive performance of the target feature, and do not evaluate the learned feature interactions quantitatively.
We propose to evaluate the quality of the graph structure by comparing edge-wise (ignoring the diagonal) the true binary adjacency matrix $A_\text{true}$ with the learned weighted \adj $A_\text{pred}=A$ with the \roc \citep{bradleyUseAreaROC1997}.
This metric reflects to what degree the feature interaction strengths of the true edges are higher than those of the true non-edges.
Ranging from zero to one, a value of $0.5$ equals a random guess. A high \roc indicates that all true edges have higher feature interaction strengths than all true non-edges.
The \roc is a `relative measure', meaning that it is not sensitive to the scale of the feature interaction strengths. This way, we are forgiving in the evaluation of the feature interactions, as we only measure if the model can distinguish between true edges and true non-edges, and not the absolute values of the feature interaction strengths.

\subsection{Pruning the feature interactions} \label{sec:pruning}
\begin{figure}[htbp]
    \centering
    \includegraphics[width=0.8\textwidth]{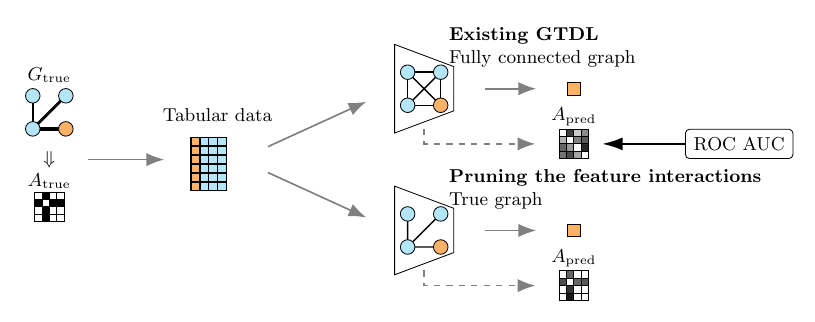}
    \caption{
        Upper branch (similar to \cref{fig:problem_statement}): Existing GTDL methods use a fully connected graph. 
        The learned adjacency matrix $A_\text{pred}$, that we extract out of \gtdl methods, is compared to the true adjacency matrix $A_\text{true}$ with the \roc to evaluate the learned feature interactions.
        Lower branch: When the graph is pruned to the true edges, \gtdl methods can only model feature interactions that are present in the true graph. By doing this, we can analyze the effect when the correct graph structure is used in \gtdl methods.
    }
    \label{fig:evaluate_interactions}
\end{figure}

To understand the effect of learning the correct graph structure, we model the \gtdl methods in two different settings, which are sketched in \cref{fig:evaluate_interactions}.
First, we train the \gtdl with a fully connected graph. This is the default setting in \gtdl methods, as the true graph in real-world datasets is not known. 
Second, we limit feature interactions to only those present in the synthetic data, effectively \textit{pruning the graph to the true edges}.
This means that the model is only allowed to learn feature interactions that are present in the true graph. Practically, this is done by masking the attention map or the graph structure within the network architecture.
This is only possible if the true graph is known, which is the case for synthetic data. By comparing the results from the fully connected and the pruned graphs, we can see how much the \gtdl methods benefit from using only the true edges.

\subsection{Setup of the experiments} \label{sec:experiment}
We conduct a standard deep learning experiment to evaluate existing \gtdl methods. That is, we optimize the prediction of the target feature $y$ given the input features $x$.
This is done with the default \gtdl setting of a fully connected graph, and with the pruned setting where the feature interactions are limited to the true edges only.
We tune the hyperparameters of the model and the learning rate, we use cross validation, and optimize the \gls{mse} with Adam \citep{kingmaAdamMethodStochastic2017}.
Further details on the splitting, training, evaluation, and \hparam tuning and cross validations can be found in \cref{sec:exp_details}.
The code is publicly available.\footnote{\url{https://github.com/elidub/gtdl_fi}}

We report the quality of the learned graph structure (\cref{sec:eval}) only for the fully connected setting, as the pruned setting trivially has perfect graph quality. The pruned graph only contains the true edges, so the learned graph structure is equal to the true graph structure.
For the predictive performance, we use the \rscore to compare the regression models.
To average over the datasets, we use the \textit{normalized} \rscore. 
This normalization is introduced by \citet{wistubaLearningHyperparameterOptimization2015} and used by \citet{feurerAutoSklearn20Handsfree2022,  grinsztajnWhyTreebasedModels2022} for instance.
Per dataset, the \rscore is normalized between zero and one, using the worst-performing and the best-performing model on that dataset as the lower and upper bound, respectively.

We run the experiments for three datasets belonging to both dataset types (\mvn and \scm), and their graph structures are shown in \cref{sec:data_gen}.
We compare all explicit \gtdl methods that have publicly published code. That is, we compare \begin{fitemize}
    \item \fignn \citep{liFiGNNModelingFeature2020},
    \item \tg \citep{yanT2GFormerOrganizingTabular2023} and
    \item \ince \citep{villaizan-valleladoGraphNeuralNetwork2024}.
\end{fitemize}
The remainder of the explicit \gtdl methods, 
\begin{fitemize}
    \item DRSA-Net \citep{zhengDeepTabularData2023},
    \item MPCFIN \citep{yeCrossFeatureInteractiveTabular2024} and
    \item Table2Graph \citep{zhouTable2GraphTransformingTabular2022},
\end{fitemize}
have not published their code repositories.
For implicit, attention-based methods, we take \ft \citep{gorishniyRevisitingDeepLearning2021} as an illustrative example. 
In \cref{sec:lit_implementation}, we discuss how these methods use and interpret the learned weighted \adj, and how we adapt the implementations to compare them.
We use the \pgm method \bdgraph \citep{mohammadiBDgraphPackageBayesian2019} as a baseline to understand how well \gtdl methods should be able to learn the feature interactions. 
Finally, we include TabPFN and XGBoost as additional baselines given their overall strong performance, to understand the difficulty of the predictive task.

\insertbib

\section{Results of structure-aware learning in GTDL} \label{sec:analysis}

To substantiate our claim that \gtdl methods should focus on learning the graph structure, we demonstrate that \gtdl methods do not accurately learn the feature interactions and that the predictive performance improves when the graph is pruned to its true edges.
Results are aggregated per dataset type, see \cref{sec:results_details} for the results per dataset.

\subsection{Feature interactions}
\begin{figure}[htbp]
    \centering
    \includegraphics[width=\textwidth]{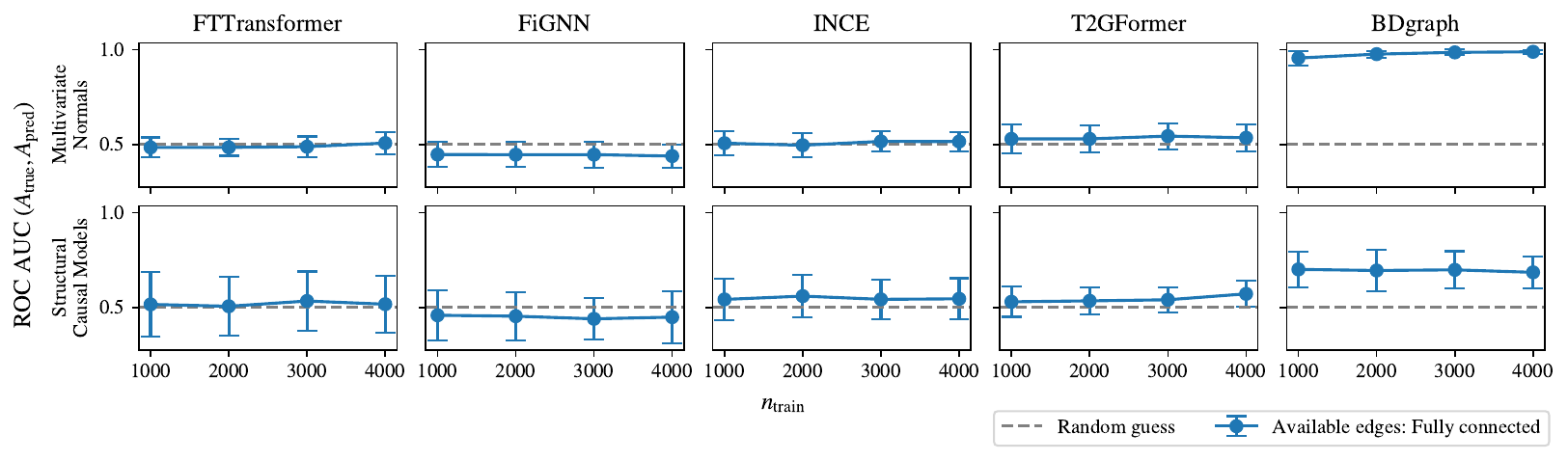}
    \caption{
    Graph quality in the form of the \roc comparing the learned weighted \adj with the true binary one, for two different dataset types.
    All \gtdl models have $\text{\roc} \approx 0.5$, which is random chance, indicating that they are not able to learn the feature interactions in any meaningful way. The \pgm method \bdgraph does learn the correct feature interactions better.
    Error bars show standard deviation across seeds, cross validations and datasets.
    }
    \label{fig:results_roc_auc}
\end{figure}

The \roc of the feature interactions is shown in \cref{fig:results_roc_auc}. For all \gtdl methods, across both datasets, the \roc is approximately $0.5$, which is equal to random chance. There is no difference in the values of the \adj whether there exists a true edge or not.
This shows that \gtdl methods do not learn an accurate graph structure. Therefore, the learned feature interactions should not be used for interpretability or explainability.
Increasing the number of training samples does not change the \roc, indicating that the poor performance of the \gtdl methods is not due to insufficient data. 
This observation holds when increasing the number of training samples up to $10^5$ samples as discussed in \cref{sec:results_ntrain}.

\Glspl{pgm}, which focus on learning the graph structure, can learn the feature interactions, while \gtdl methods cannot. 
The \pgm method \bdgraph has an \roc very close to one for the \mvn datasets. 
Even for the \scm datasets, which have nonlinear feature interactions, \bdgraph can achieve reasonable \roc values. 
The fact that this \pgm method, a non-deep learning method, can learn the feature interactions, while these advanced \gtdl methods cannot, 
suggests that \gtdl methods have room for improvement in learning the graph structure.

\subsection{Predictive performance}
\begin{figure}[htbp]
    \centering
    \includegraphics[width=\textwidth]{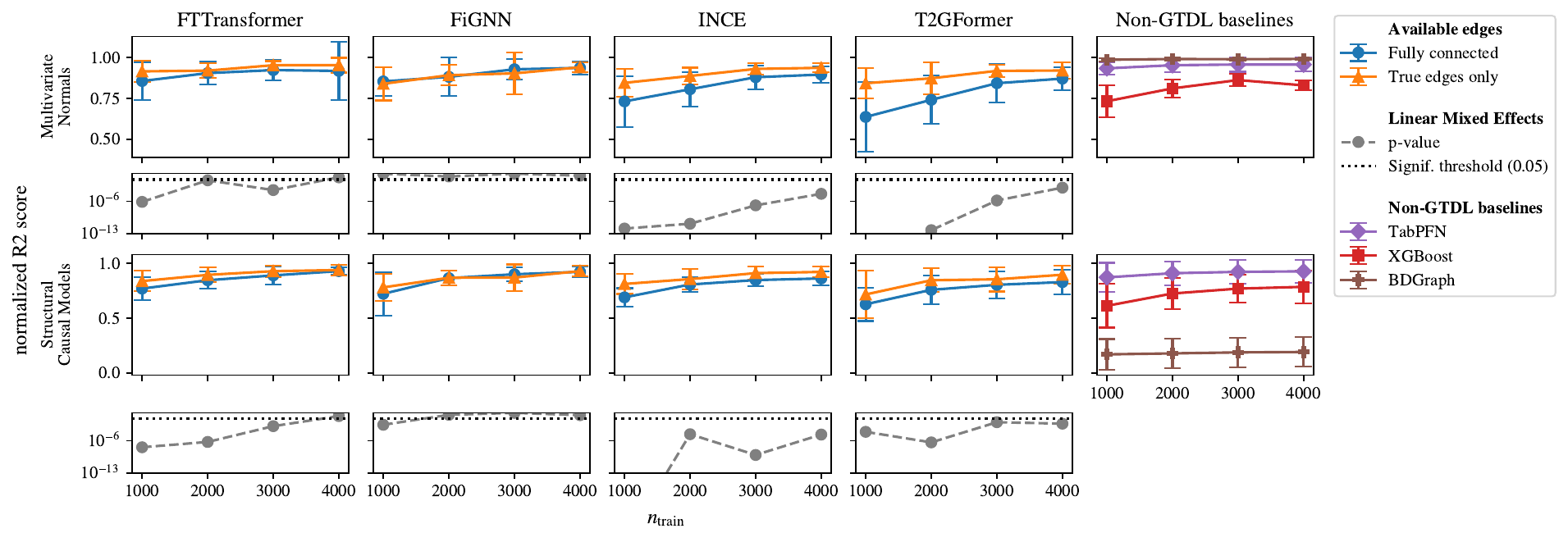}
    \caption{Predictive performance for two different dataset types.
    When the graph is pruned to its true edges, the predictive performance is, in most cases, better compared to the fully connected graph. The difference reduces as the number of training samples increases.
    This is confirmed by the $p$-values of the Linear Mixed-Effects Model, where lower $p$-values indicate stronger evidence that the pruned graph results are better than the fully connected graph results.
    Error bars show standard deviation across seeds, cross validations and datasets.
    }
    \label{fig:results_r2score}
\end{figure}

The \rscore of the prediction of the target feature is shown in \cref{fig:results_r2score}. The key takeaway is that, in general, pruning the graph to the true edges improves the predictive performance of \gtdl methods. This result indicates the importance of incorporating accurate structural information into \gtdl models. When the graph is pruned to only include true edges, the models are less likely to overfit to spurious or irrelevant feature interactions. 
Restricting the model to only the true interactions simplifies the optimization landscape, allowing the learning algorithm to focus on meaningful relationships rather than being distracted by false edges. This leads to better generalization and higher predictive accuracy.
In contrast, fully-connected models must learn to ignore many false edges, which can introduce noise and make optimization more difficult, especially when data is limited.
This finding suggests that the inability of current \gtdl methods to recover the true graph structure (as shown by the \roc results) is not just a theoretical issue, but has practical consequences for predictive performance. If the true graph is known or can be estimated reliably, enforcing this structure can provide a boost in performance.

We use the Linear Mixed-Effects Model \citep{lindstromNewtonRaphsonEM1988,pinheiro2000mixed} to assess whether the pruned graph outperforms the fully connected graph. We elaborate on the statistical testing procedure in \cref{sec:statistical_test}. 
From \cref{fig:results_r2score}, we observe that, except for \fignn, the pruned graph results are both visually (first and third row) and statistically significantly (second and fourth row) better than the fully connected graph results for low number of training samples.

In \cref{sec:node-level_graph-level}, we discuss results that indicate that this is because \fignn treats the task of predicting the target feature as a graph-level task, while all other models treat it as a node-level task.

Regarding baselines, the \pgm method \bdgraph performs, as expected, well on the \mvn datasets, and poorly on the \scm datasets. The \scm datasets have nonlinear feature interactions, while \bdgraph can only model linear feature interactions. In most cases, \gtdl methods outperform XGBoost, while TabPFN outperforms \gtdl methods.
A possible explanation for the performance gap is that GTDL and TabPFN may better exploit the tabular structure of data through learned representations, whereas XGBoost relies on an ensemble of weak learners that may not fully capture feature interactions inherent in such structures. This underlines the need for deep learning methods due to their flexibility to learn nonlinear relationships. 

Furthermore, the benefit of incorporating the true graph increases by reducing the number of training samples. When ample data is available, models benefit less from incorporating the graph structure correctly, but when data is scarce, leveraging the graph structure improves the predictions. 
This is in line with the general notion of geometric deep learning, where symmetries in the data are used to improve the learning process \citep{bronsteinGeometricDeepLearning2021}. 
When data is scarce, explicitly incorporating the symmetries from the data into the model is beneficial. However, an abundance of data facilitates an implicit learning of these symmetries \citep{marchettiHarmonicsLearningUniversal2024}, reducing the benefit of explicitly incorporating them.

This observation holds when increasing the number of training samples up to $10^5$ samples, discussed in \cref{sec:results_ntrain}.

\insertbib

\glsreset{pgm}
\glsreset{gtdl}
\section{Conclusion, limitations, and future work}\label{sec:conclusion}

In this work, we have analyzed the capability of \gtdl methods to learn feature interactions in tabular data.
Inspired by the principles of \pgms, we proposed to use synthetic tabular datasets with known ground-truth graph structures, enabling the \gtdl community to quantitatively assess whether models accurately capture the intended graph structure.
Current \gtdl approaches often produce graph structures used for interpretation, yet our analysis shows that these structures fail to reflect the true interactions among features. This indicates that the mechanisms of message-passing in \gnns, and attention in transformers, does not work as intended for tabular data.
Our empirical findings demonstrate that when models operate on accurate interaction structures, predictive performance improves, highlighting that structural fidelity is not merely a matter of explainability, but a core driver of performance.

A potential risk could be the interpretation of the attention map as a weighted adjacency matrix.
First, the attention mechanism is not designed to specifically model feature interactions or for explainability. If attention can be used for explainability is an ongoing debate \citep{bibalAttentionExplanationIntroduction2022,lopardoAttentionMeetsPosthoc2024}, although this debate in the literature mainly focuses around natural language tasks.
However, our interpretation for tabular data—that non-interacting features will exhibit low attention values while interacting features will manifest higher attention—is deliberately modest and pragmatic, rather than relying on attention as a fully explanatory tool.
Second, while we acknowledge that aggregating attention across layers and heads might obscure certain signals, we find no strong evidence that alternative aggregation strategies would significantly improve graph recovery or alter the overall conclusions.

We highlight three directions how the analysis of this work can be extended in future work.
First, 
future work should move beyond evaluating the learned graph structure (i.e., the presence of edges), but also consider the functional form of the feature interactions (i.e., the type of edges). Learning \textit{how} features interact, and \textit{in what way}, allows for more nuanced, robust, and interpretable modeling of feature relationships.
Second,
the evaluated datasets and graphs could be more expressive and challenging.
Examples include larger graphs with richer topology, missing nodes, and more complex feature interactions involving categorical features, as well as real-world datasets with known ground-truth structure (e.g., knowledge graphs).
However, our results showed that \gtdl methods struggle to learn relatively simple, small graph structures, which suggests that improving robustness and structure induction on basic cases remains a priority before scaling to such settings.
Finally, 
structure-aware modeling should be extended beyond flat tables to richer data modalities. Time-series data \citep{padhiTabularTransformersModeling2021} and relational databases \citep{feyRelationalDeepLearning2023,robinsonRelBenchBenchmarkDeep2024,congObservatoryCharacterizingEmbeddings2024,dwivediRelationalGraphTransformer2025} present new challenges for learning and validating feature interactions over time or across relational contexts.
\Gls{rdl} \citep{feyRelationalDeepLearning2023} could ask the same question as we propose \gtdl should do: how do columns/features within a table relate? Currently, table-row embeddings in \gls{rdl} lack structural bias from the graph structure of columns. Furthermore, our approach could be extended from table-level to relational database-level. When doing node-level prediction, does \gls{rdl} rely on meaningful primary-foreign relationships?

Future work should build on the insights of this work to develop \gtdl methods that more effectively learn and leverage feature interactions in tabular data. By prioritizing structural fidelity alongside predictive accuracy, future \gtdl models can unlock their full potential.

\insertbib

\insertbibmain

\appendix

\section{Implementations from literature}\label{sec:lit_implementation}

This section contains implementation details of the \gtdl models that are evaluated in \cref{sec:analysis}. In the following \cref{sec:tg_details,sec:ince_details,sec:ft_details,sec:fignn_details}, each first paragraph explains what kind of graph is learned, and where and how this is used by the original authors. In the subsections, we discuss the adaptations we made to make the learned graph structures compatible with our discussion and implementation. 

\paragraph{Additional notation.}
We use $N$ for the number of samples, $L$ as the number of layers in the network, $H$ as the number of transformer heads, and $p$ as the number of features. 
To indicate values that are ranging between zero and one, we denote the corresponding set as $\RRone$. 

\paragraph{Interpreting the attention map as a weighted adjacency matrix.}
For most methods (\ft, \tg and \fignn), the learned graph comes from averaging the attention map $a \in \RRone^{N \times L \times H \times p \times p}$ over the samples, layers, and heads. This attention map is normalized with softmax across the last dimension. We note individual values from the attention map as $a_{ilhjk}$, where $i$ is the sample index, $l$ is the layer index, $h$ is the head index, and $j,k$ are the feature indices. So the average attention map is $a_{jk} = \frac{1}{N \times L \times H} \sum_{ilh} a_{ilhjk} \in \RRone^{p \times p}$.

We want to interpret the average attention map $a_{jk}$ as the weighted adjacency matrix $A_{jk}$. For this, we have to `denormalize' the average attention map. As the attention map $a$ is normalized with a softmax, the last dimension (the rows in the attention map per individual layer and head) sum to one, such that $\sum_k a_{ilhjk} = 1$ for all $i,l,h,j$. This gives a problem, as the maximum value the attention map $a_{ilhjk}$ can have cannot have two values close to $1$ in the same row, while the weighted adjacency matrix $A_{jk}$ should be able to have multiple values close to $1$ in the same row.

To `denormalize' the attention map, we add two steps. First, we set the diagonal of the attention map to zero, as the self-interactions should not be taken into account during evaluation of the feature interactions:
\begin{equation*}
  a_{ilhjk} = 0 \quad \forall j = k.
\end{equation*}
Second, we divide the attention maps by the maximum value across the row to obtain the adjacency matrices:
\begin{equation*}
  A_{ilhjk} = a_{ilhjk} / \max_k(a_{ilhjk}).
\end{equation*}
By doing this, all the values with the highest attention across that row now have a value of $1$ in the weighted adjacency matrix. Ignoring the diagonal in the attention map is a key step in this procedure: If the model learns that it should not give high attention to the non-diagonal values (as those features are not related), the model should learn to give high attention to the diagonal values. The diagonal values are excluded in the denormalization and do not affect the adjacency matrix.

\subsection{FT-Transformer} \label{sec:ft_details}

\ft \citep{gorishniyRevisitingDeepLearning2021} (\url{https://github.com/yandex-research/rtdl-revisiting-models}) learns the attention map $a \in \RRone^{N \times L \times H \times p \times p}$. 
\citep{gorishniyRevisitingDeepLearning2021} interpret in Section~5.3 the average attention map $a_{jk} = \frac{1}{N \times L \times H} \sum_{ilh} a_{ilhjk} \in \RRone^{p \times p}$ as feature importance. For a few real-world datasets, they compare it to Integrated Gradients \citep{sundararajanAxiomaticAttributionDeep2017} using rank correlation and find that it performs similarly.

As the attention map is normalized with the softmax, we denormalize it to obtain the weighted adjacency matrix $A$ as described above.
This is the only post-hoc adaptation we made to the original implementation.

\FloatBarrier

\subsection{\tg} \label{sec:tg_details}
\tg \citep{yanT2GFormerOrganizingTabular2023} (\url{https://github.com/jyansir/t2g-former}) learn a \frgraph $a \in \RRone^{N \times L \times H \times p \times p }$ (Equation 6 in \citep{yanT2GFormerOrganizingTabular2023}). 
The strength of the graph can be interpreted as the strength of the relations between the features. 
Section~5.3 and Figure~3 in \citep{yanT2GFormerOrganizingTabular2023} show the \frgraph for two real-world datasets. 

Instead of the Hadamard product in equation 6 in \citep{yanT2GFormerOrganizingTabular2023} to construct the \frgraph, we use a sum, consistent with their \href{https://github.com/jyansir/t2g-former/blob/3ad093d7967b726b265606251ec267daa0756427/bin/t2g_former.py#L239}{code implementation of the \frgraph}. 
\citep{yanT2GFormerOrganizingTabular2023} present the \frgraph from the first and the last layer of the network. We assume that these are averaged over the samples and the heads. Instead, we average over all the layers, following the approach of \ft.
As the \frgraph is normalized with the softmax, we denormalize the \frgraph to obtain the weighted adjacency matrix $A$ as described above.
\FloatBarrier
    
\subsection{INCE} \label{sec:ince_details}
\ince \citep{villaizan-valleladoGraphNeuralNetwork2024} (\url{https://github.com/MatteoSalvatori/INCE}) learn edge embeddings $e \in \RR^{N \times (p (p-1)) \times d_\text{emb}}$ with $d_\text{emb}$ the embedding dimension and $(p (p-1))$ the number of edges in a fully connected graph excluding self-loops. 
Section~6.2 of \citep{villaizan-valleladoGraphNeuralNetwork2024} presents an algorithm to calculate the feature-feature interaction $p_\text{int} \in \RRone^{p \times p}$ from the edge embeddings. 
Figure~11 in \citep{villaizan-valleladoGraphNeuralNetwork2024} shows the feature-feature interaction $p_\text{int}$ on a real-world dataset. 

A lower value of $p_\text{int}$ implies more significance. Therefore, we apply one additional step to obtain the weighted adjacency matrix $A = 1 - p_\text{int}$. 

\subsection{FiGNN} \label{sec:fignn_details}
\fignn \citep{liFiGNNModelingFeature2020} (\url{https://github.com/CRIPAC-DIG/Fi_GNN/tree/7e207b2ffb4f25b63d2079cf7761d09e5dedf6e8}\footnote{This on purpose a specific commit, as there are different implementations.}) learn a feature graph in the form of attentional edge weights $a \in \RRone^{N \times (p-1) \times (p-1)}$ for the input features (equations~4 and 5 in \citet{liFiGNNModelingFeature2020}). 
The edge weights are interpreted as the importance of the interactions. Therefore, they are used to providing explanations on the relationship between different features. In Section~4.5 and Figure~5, the edge weights are presented as a heat map, and are used to explain the relations between features on a real-world dataset.

As the attention edge weights are normalized with the softmax, we denormalize them to obtain the weighted adjacency matrix $A$ as described above. 
The code implementation of \fignn is published with TensorFlow. As our implementation is in PyTorch, we have adapted the code to PyTorch.
The learned adjacency matrix is of size $(p-1) \times (p-1)$, we impute an additional row and column for the target feature with values of zero. 

There are different implementation versions of \fignn.
The first version of \fignn has been presented at CIKM in November 2019, which is identical to version~1 on ArXiv \citep{liFiGNN_ArXiv_v1}. We use the implementation of this version. In July 2020, version~2 on ArXiv \citep{liFiGNN_ArXiv_v2} was published, and the main repository (\url{https://github.com/CRIPAC-DIG/Fi_GNN/}) was updated accordingly. Version~2 has some additional attention layers. Furthermore, when inspecting the published code. We observed that this second version does not have a trainable feature graph in its code implementation. Therefore, we stick to the original code implementation of version~1 (\url{https://github.com/CRIPAC-DIG/Fi_GNN/tree/7e207b2ffb4f25b63d2079cf7761d09e5dedf6e8}). 

\FloatBarrier

\insertbib

\section{Data generation}\label{sec:data_gen}

In this section, we describe the graph and data generation process of the two synthetic dataset approaches introduced in \cref{sec:dataset} and their \hparams used in \cref{sec:analysis}.

\paragraph{Multivariate normals.}

We follow the default procedure of generating conditional multivariate data, \citep{mohammadiBayesianStructureLearning2015}:
\begin{enumerate}[label=(\roman*)]
    \item Sample a true graph structure $G_\text{true} \in \mathbb{R}^{p \times p}$ from the Bernoulli distribution with an edge inclusion probability $P_\text{edge}$.
    \item Sample a covariance matrix $\Sigma_G \RRpp$ from the G-Wishart distribution \citep{roveratoHyperInverseWishart2002, letacWishartDistributionsDecomposable2007}, which is conditioned on the graph structure $G_\text{true}$;\footnote{In fact, we sample the precision matrix $K_G = \Sigma_G^{-1}$ form the G-Wishart distribution, and invert it to obtain the covariance matrix $\Sigma_G$. The underscore $\cdot_G$ indicates that the matrix is conditioned on the graph structure $G_\text{true}$.}
    \item Obtain $n$ samples $D \in \RR^{n \times p}$ from a multivariate normal distribution $\mathcal{N}(0, \Sigma_G)$.
\end{enumerate}

\begin{figure}[htbp]
    \centering
    \includegraphics[width=0.32\textwidth]{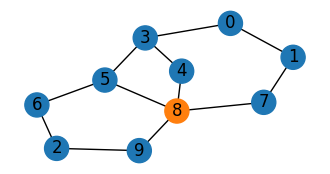}
    \hfill
    \includegraphics[width=0.32\textwidth]{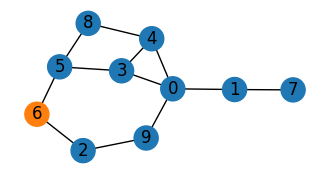}
    \hfill
    \includegraphics[width=0.32\textwidth]{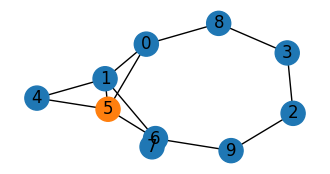}
    \caption{Graphs used in experiments for the MVN1, MVN2 and MVN3 datasets.}
    \label{fig:mvns}
\end{figure}

In our experiments, we have $p = 10$ nodes, and an edge inclusion probability of $P_\text{edge} = 0.267$. This results in the graph structures as depicted in \cref{fig:mvns}. 

\FloatBarrier

\paragraph{Structural causal models.}
We follow a similar setup as \citep{hollmannAccuratePredictionsSmall2025} to generate an \scm and sample data conditional on the graph. They show that with their setup, the synthesized data is similar to real-world tabular data. 

\begin{figure}[htbp]
    \centering
    \includegraphics[width=0.32\textwidth]{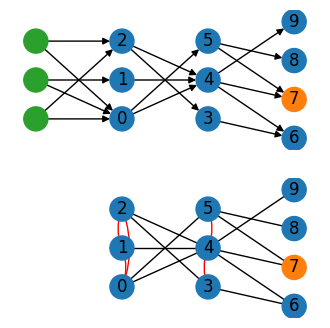}
    \hfill
    \includegraphics[width=0.32\textwidth]{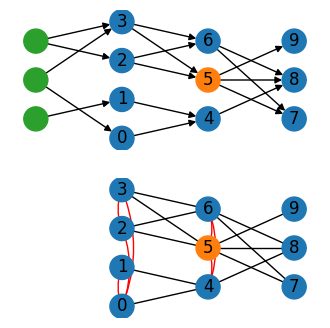}
    \hfill
    \includegraphics[width=0.32\textwidth]{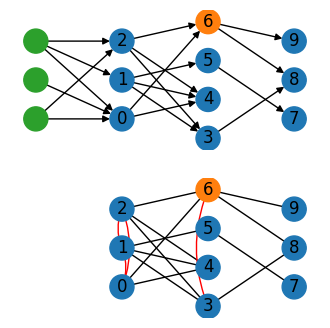}
    \caption{Top: \glspl{dag} used in experiments for the SCM1, SCM2 and SCM3 datasets. Bottom: Corresponding undirected graph structures $G_\text{true}$ after moralization and marginalization. Moralized and marginalized edges are depicted in red.}
    \label{fig:scms}
\end{figure}

\begin{enumerate}[label=(\roman*)]
    \item Randomly sample a \gls{dag}, with $n_\text{root}$ root nodes and $p$ child nodes with a probability of $P_\text{edge}$ of an incoming edge.
    The undirected graph structure $G_\text{true}$ is obtained by moralizing and marginalizing the \gls{dag} \citep{cowellProbabilisticNetworksExpert1999}. Moralizing makes the graph undirected by dropping the direction of the existing edges and connecting all parents of a child node. Marginalizing removes the root nodes from the graph, as they are not part of the dataset $D$. When marginalizing a node, we connect all neighbors of the removed node to each other. In this work, only the root nodes are marginalized. This makes the order of moralization and marginalization irrelevant. The red lines in \cref{fig:scms} show the new edges that are added by moralization and marginalization.
    \item Randomly sample deterministic computational mappings $f_i$ for each child node $i$ in the graph, where the mappings are smooth nonlinear functions, randomly picked from the set of maps listed in \cref{tab:computational_maps}.
    A computational map defines how a child node $i$ is computed from its parents. They take all the values of the incoming edges as input, and the output is the value of the child node $i$.
    \item Randomly sample root nodes $x_\text{root} \sim \mathcal{N}(0, 1) \in \RR^{n \times n_\text{root}}$ and 
    traverse the \gls{dag} in a topological order, $x_i = f_i(x_{i, \text{parents}}) \in \RR^{n \times 1}$. Each output $x_i$ is normalized, clipped between $(-3, 3)$, and Gaussian noise $\mathcal{N}(0, 0.5)$ is added. This is summarized by
    \begin{equation}
        x_i = \text{clip}(\text{normalize}(f_i(x_{i, \text{parents}})) + \mathcal{N}(0, 0.5), -3, 3).
    \end{equation}
    We consider all the traversed outputs $x_i$ as the dataset $D \in \RR^{n \times p}$. 
\end{enumerate}

Each \gls{dag} has $p = 10$ nodes. These $p$ nodes are evenly distributed over $n_\text{DAG layers} = 3$ layers, where each layer has a minimum of $3$ nodes. The \gls{dag} has a `zeroth' layer of $n_\text{root} = 3$ root nodes. 
This means that each layer has $3$ or $4$ nodes.
Each node has a $P_\text{edge} = 0.5$ probability of having an edge to the nodes in the next layer. With these \hparams, the three \glspl{dag} that are used in \cref{sec:analysis} are shown in \cref{fig:scms}.

\begin{table}[htbp]
\caption{Computational maps}
\label{tab:computational_maps}
\centering
\begin{tabular}{ll}
    \hline
    \# parents & $f(x_\text{parents})$ \\
    \hline
    1 & $x_1^2 / 3$ \\
    & $0{,}5\, x_1^2 + 3\, x_1$ \\
    & $-|x_1| + 4\, x_1$ \\
    \hline
    2 & $(x_1 x_2 + x_1^2) / 2$ \\
    & $x_1^2 + x_2^2 - x_1 x_2$ \\
    & $-(x_1 + x_2)^2 + x_1 x_2$ \\
    \hline
    3 & $(x_1 x_2 + x_3^2) / 3$ \\
    & $-x_1^2 + x_2 x_3 + x_3$ \\
    & $(x_1 + x_2 + x_3) + x_1 x_3$ \\
    \hline
\end{tabular}
\end{table}

\FloatBarrier

\insertbib

\section{Experiment details}\label{sec:exp_details}

\paragraph{Data splitting.}

We adapt our train, validation, and test splitting and our tuning strategy to balance between a fair comparison between different dataset sizes and an efficient \hparam tuning.

Following \citep{grinsztajnWhyTreebasedModels2022}, we differentiate between a validation set used for early stopping \Dvales and a validation set used for \hparam tuning \Dvalhparam, such that we have four disjoint sets: \Dtrain, \Dvales, \Dvalhparam, and \Dtest.
We vary the number of training samples \ntrain in our experiments between $1000$ and $4000$, and set both $\ntest = \nvalhparam = 2500$ and $\nvales = 0.25 \ntrain$.

\begin{figure}[htbp]
    \centering
    \includegraphics[width = 0.9 \linewidth]{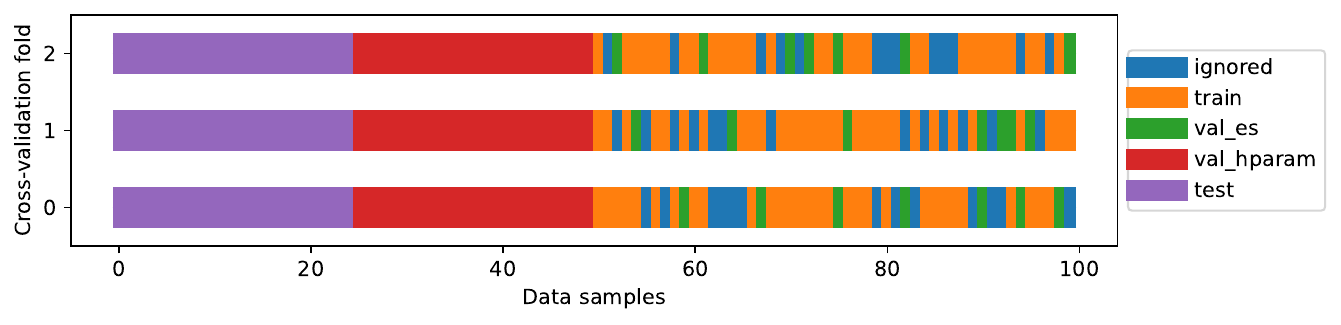}
    \caption{Splitting strategy for the example $\nsamples = 100$ and $\ntrain = 0.3 \nsamples$.
    The training set and the validation set for early stopping are randomly sampled for each fold. 
    The test set and validation set for \hparam tuning are fixed.
    }
    \label{fig:split_example}
\end{figure}

In our experiments, we do not change \Dvalhparam and \Dtest to limit the number of cross-validation and iterations we have to do. 
We randomly sample \Dtrain and \Dvales for each fold. 
This strategy is visualized in \cref{fig:split_example}.
For $\ntrain = 1000$ samples we evaluate over $4$ folds, $\ntrain = 2000$ over $3$ fold, for $\ntrain = 3000$ over $2$ folds and for $\ntrain = 4000$ over $1$ fold.

\paragraph{Training and evaluation.}
We minimize the \mse loss function and optimize using Adam \citep{kingmaAdamMethodStochastic2017} with a fixed batch size of $256$ and tune the learning rate together with the other model \hparams. 
We continue training until the validation loss does not improve for $10$ epochs. There is a theoretical upper bound of $400$ epochs, which is rarely reached in practice.
We select the best \hparams that minimize the \mse on the separate \hparam validation set. After tuning, we run $10$ runs per cross-validation fold. We report the \rscore to evaluate the predictive performance of the target feature, and the \roc to evaluate the learned feature interactions.

\paragraph{Hyperparameter tuning.}

For every combination of network, dataset, and \ntrain, we tune the model's \hparams and the learning rate. For all models, we use \gls{tpe} \citep{bergstraAlgorithmsHyperParameterOptimization2011}, a Bayesian optimization technique within the Optuna library \citep{akibaOptunaNextgenerationHyperparameter2019}.
We run a total of $50$ trials for each setting, where the first trial has the default \hparams of the implementation. We keep the default setting of the Optuna implementation, where the first $10$ trials are done with random search. 
The hyperparameters of TabPFN are not tuned, as it is a pretrained model that can get accurate predictions without fine-tuning \citep{hollmannAccuratePredictionsSmall2025}.

The \hparam distribution and the default \hparams of all models are listed in \cref{tab:ft_hparam,tab:tg_hparam,tab:ince_hparam,tab:fignn_hparam,tab:xgb_hparam}. For all models, the search space and default values are taken from the original implementations if not specified otherwise in the caption of the tables. The search space of layer count and embedding size is set the same for fairer comparison across models. The distribution space of the learning rate is $\loguniform[\tpn{-5}, \tpn{-3}]$ with a default value of $\tpn{-3}$ for all models.

\FloatBarrier

\begin{table}[H]
  \caption{\ft \citep{gorishniyRevisitingDeepLearning2021} \hparam space.}
  \label{tab:ft_hparam}
  \centering
  \begin{tabular}{lll}
    \toprule
    Parameter & Distribution & Default \\
    \midrule
    Layer count & $\uniformint[1, 6]$ & $3$ \\
    Embedding size & $\{8, 16, 32, 64, 128, 264\}$ & $128$ \\
    Attention head count & - & $8$ \\
    Attention dropout & $\uniform[0.0, 0.5]$ & $0.2$ \\
    FFN size factor & $\uniform[\sfrac{2}{3}, \sfrac{7}{3}]$ & \sfrac{4}{3} \\
    FFN dropout & $\uniform[0.0, 0.5]$ & $0.1$ \\
    Residual dropout & $\uniform[0.0, 0.2]$ & $0$ \\
    \bottomrule
  \end{tabular}
\end{table}

\begin{table}[H]
  \caption{\tg \citep{yanT2GFormerOrganizingTabular2023} \hparam space. Default values are taken from the same as from \ft.}
  \label{tab:tg_hparam}
  \centering
  \begin{tabular}{lll}
    \toprule
    Parameter & Distribution & Default \\
    \midrule
    Layer count & $\uniformint[1, 6]$ & $3$ \\
    Embedding size & $\{8, 16, 32, 64, 128, 264\}$ & $128$ \\
    Attention head count & - & $8$ \\
    Attention dropout & $\uniform[0.0, 0.5]$ & $0.2$ \\
    FFN size factor& $\uniform[\sfrac{2}{3}, \sfrac{7}{3}]$ & \sfrac{4}{3} \\
    FFN dropout & $\uniform[0.0, 0.5]$ & $0.1$ \\
    \bottomrule
  \end{tabular}
\end{table}

\begin{table}[H]
  \caption{\ince \citep{villaizan-valleladoGraphNeuralNetwork2024} \hparam space.}
  \label{tab:ince_hparam}
  \centering
  \begin{tabular}{lll}
    \toprule
    Parameter & Distribution & Default \\
    \midrule
    Layer count & $\uniformint[1, 6]$ & $4$ \\
    Embedding size & $\{8, 16, 32, 64, 128, 264\}$ & $128$ \\
    MLP layer count & $\{1, 2, 3, 4\}$ & $3$ \\
    Dropout & $\uniform[0.0, 0.5]$ & $0$ \\
    \bottomrule
  \end{tabular}
\end{table}

\begin{table}[H]
  \caption{\fignn \citep{liFiGNNModelingFeature2020} \hparam space. The distribution space was not shared by the original implementation.}
  \label{tab:fignn_hparam}
  \centering
  \begin{tabular}{lll}
    \toprule
    Parameter & Distribution & Default \\
    \midrule
    Layer count & $\uniformint[1, 6]$ & $3$ \\
    Embedding size & $\{8, 16, 32, 64, 128, 264\}$ & $16$ \\
    Dropout & $\uniform[0.0, 0.5]$ & $0$ \\
    \bottomrule
  \end{tabular}
\end{table}

\begin{table}[H]
  \caption{XGBoost \citep{chenXGBoostScalableTree2016} \hparam space. Default values are taken from the official implementation \citep{chenXgboostExtremeGradient2026}, distribution space is the same as in \citep{grinsztajnWhyTreebasedModels2022}.}
  \label{tab:xgb_hparam}
  \centering
  \begin{tabular}{lll}
    \toprule
    Parameter & Distribution & Default \\
    \midrule
    Max depth & $\uniformint[3, 10]$ & $6$ \\
    Min child weight & $\loguniform[10^{-8}, 10^5]$ & $1$ \\
    Subsample & $\uniform[0.5, 1.0]$ & $1.0$ \\
    Learning rate (eta) & $\loguniform[10^{-5}, 1.0]$ & $0.3$ \\
    Colsample by level & $\uniform[0.5, 1.0]$ & $1.0$ \\
    Colsample by tree & $\uniform[0.5, 1.0]$ & $1.0$ \\
    Gamma & $\loguniform[10^{-8}, 100]$ & $10^{-8}$ \\
    Lambda & $\loguniform[10^{-8}, 100]$ & $1.0$ \\
    Alpha & $\loguniform[10^{-8}, 100]$ & $10^{-8}$ \\
    \bottomrule
  \end{tabular}
\end{table}

\FloatBarrier

\subsection{Statistical testing} \label{sec:statistical_test}
To assess the statistical significance of the predictive performance improvement when using the pruned graph compared to the fully connected graph, we used a Linear Mixed Effects Model \citep{lindstromNewtonRaphsonEM1988,pinheiro2000mixed}.
We selected this approach over aggregating results per fold (e.g., for a Wilcoxon signed-rank test \citep{wilcoxonIndividualComparisonsRanking1945}) to avoid the loss of statistical power, given the limited number of cross-validation folds $(4, 3, 2, 1)$ available for $\ntrain = (1000, 2000, 3000, 4000)$, respectively.
Since we evaluate 10 random seeds per fold, the data exhibits a hierarchical structure. Treating these seeds as independent samples would violate the independence assumption of standard tests and lead to pseudoreplication \citep{nadeauInferenceGeneralizationError1999}.
Therefore, we modeled the cross-validation fold as a random effect to account for the correlation between seeds, and the graph type (pruned or fully connected) as a fixed effect.
Although not documented in this work, we compared the Linear Mixed Effects Model with the one-sided Mann-Whitney U test \citep{mannTestWhetherOne1947} and found comparable $p$-values.

\insertbib

\section{Additional results} \label{sec:results_details}

\subsection{Results per dataset}

\begin{figure}[htbp]
    \centering
    \includegraphics[width=\textwidth]{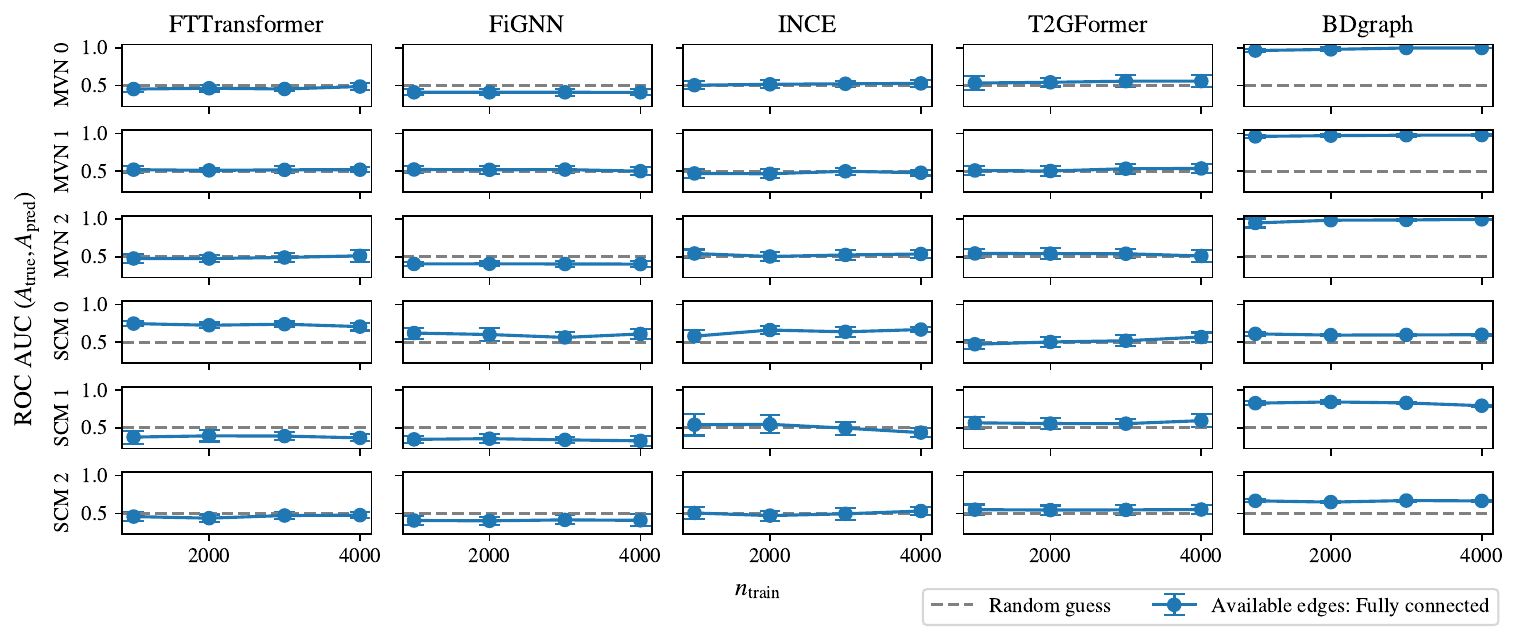}
    \caption{
    Graph quality in the form of the \roc comparing the learned weighted \adj with the true binary one, for six different datasets.
    Most \gtdl models have $\text{\roc} \approx 0.5$, which is random chance, indicating that they are not able to learn the feature interactions in any meaningful way. The \pgm method \bdgraph can learn the feature interactions.
    Error bars show standard deviation across seeds and cross validations.
    See \cref{fig:results_roc_auc} for the results aggregated over the two dataset types.
    }
    \label{fig:results_roc_auc_all}
\end{figure}

\begin{figure}[htbp]
    \centering
    \includegraphics[width=\textwidth]{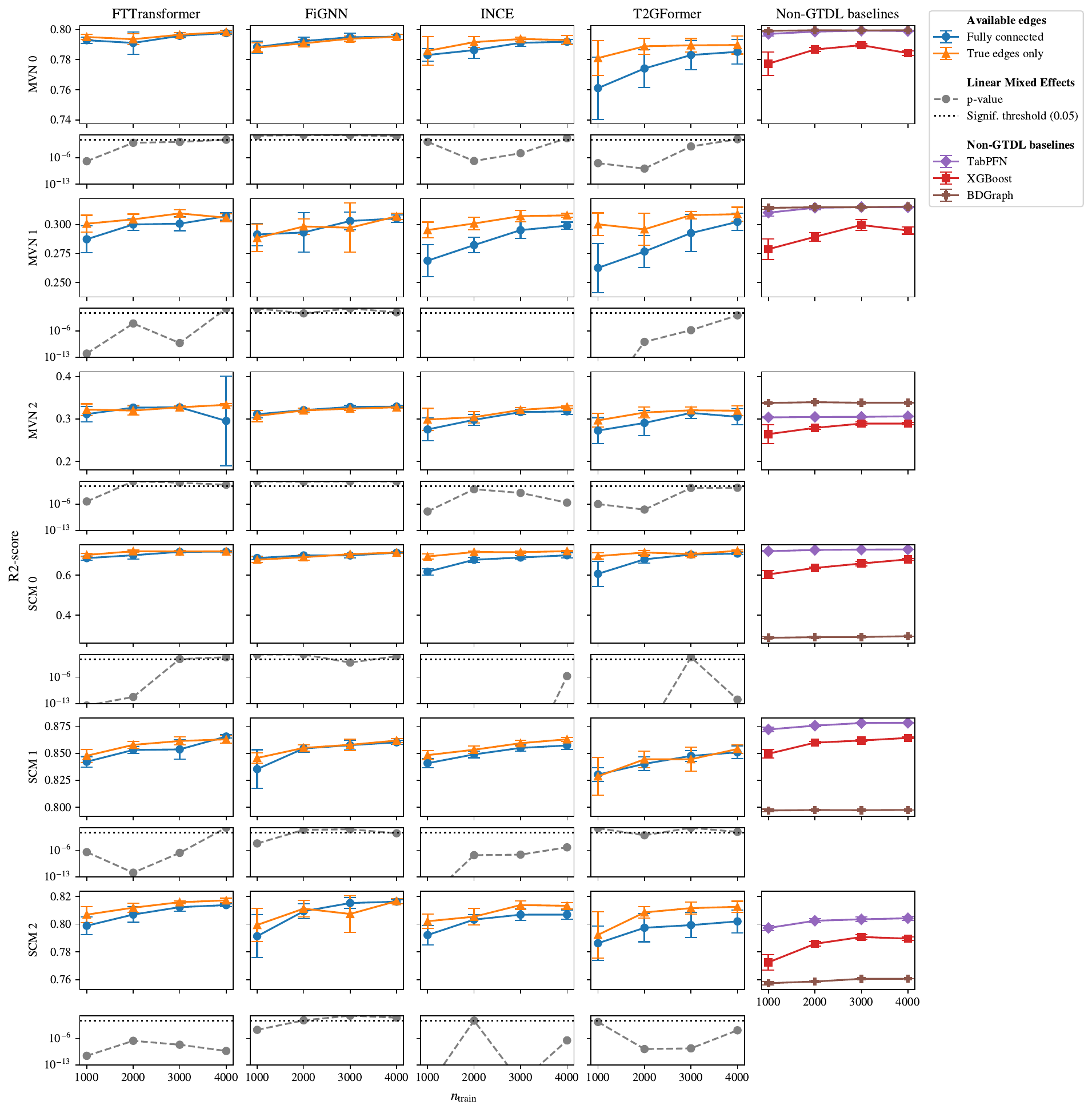}
    \caption{
    Predictive performance for six different datasets.
    When the graph is pruned to its true edges, the predictive performance is, in most cases, better compared to the fully connected graph. The difference reduces as the number of training samples increases.
    This is confirmed by the $p$-values of the Linear Mixed-Effects Model, where lower $p$-values indicate stronger evidence that the pruned graph results are better than the fully connected graph results. When $p$-values are not visible, they are below the lower limit of $10^{-13}$.
    Error bars show standard deviation across seeds and cross validations.
    See \cref{fig:results_r2score} for the results aggregated over the two dataset types.
    }
    \label{fig:results_r2score_all}
\end{figure}

In \cref{sec:analysis} we have discussed the results of the learned graph structure and the predictive performance of the target feature while aggregating over three datasets per the dataset type \mvn and \scm. 
In \cref{fig:results_roc_auc_all,tab:results_rocauc} we show the results per individual dataset for the learned graph structure, in \cref{fig:results_r2score_all,tab:results_r2score} we show them for the predictive performance. The results are consistent with the results shown in \cref{fig:results_roc_auc} and \cref{fig:results_r2score}; no new insights are gained.

\begin{landscape}
\begin{longtable}{llrrrrrrrrrr}
\caption{
    Graph quality in the form of the ROC AUC comparing the learned weighted adjacency matrix with the true binary one, for six different datasets and five different models.
    The statistics $\mu$ and $\sigma$ represent the mean and standard deviation aggregated over seeds and cross validations.
    See \cref{fig:results_roc_auc_all} for the same results.
} \label{tab:results_rocauc} \\
\toprule
 & Model & \multicolumn{2}{l}{FTTransformer} & \multicolumn{2}{l}{FiGNN} & \multicolumn{2}{l}{INCE} & \multicolumn{2}{l}{T2GFormer} & \multicolumn{2}{l}{BDgraph} \\
 & Graph & \multicolumn{2}{l}{Fully connected} & \multicolumn{2}{l}{Fully connected} & \multicolumn{2}{l}{Fully connected} & \multicolumn{2}{l}{Fully connected} & \multicolumn{2}{l}{Fully connected} \\
 & Statistic & $\mu$ & $\sigma$ & $\mu$ & $\sigma$ & $\mu$ & $\sigma$ & $\mu$ & $\sigma$ & $\mu$ & $\sigma$ \\
Dataset & $n_{\mathrm{train}}$ &  &  &  &  &  &  &  &  &  &  \\
\midrule
\endfirsthead
\caption[]{
    Graph quality in the form of the ROC AUC comparing the learned weighted adjacency matrix with the true binary one, for six different datasets and five different models.
    The statistics $\mu$ and $\sigma$ represent the mean and standard deviation aggregated over seeds and cross validations.
    See \cref{fig:results_roc_auc_all} for the same results.
} \\
\toprule
 & Model & \multicolumn{2}{l}{FTTransformer} & \multicolumn{2}{l}{FiGNN} & \multicolumn{2}{l}{INCE} & \multicolumn{2}{l}{T2GFormer} & \multicolumn{2}{l}{BDgraph} \\
 & Graph & \multicolumn{2}{l}{Fully connected} & \multicolumn{2}{l}{Fully connected} & \multicolumn{2}{l}{Fully connected} & \multicolumn{2}{l}{Fully connected} & \multicolumn{2}{l}{Fully connected} \\
 & Statistic & $\mu$ & $\sigma$ & $\mu$ & $\sigma$ & $\mu$ & $\sigma$ & $\mu$ & $\sigma$ & $\mu$ & $\sigma$ \\
Dataset & $n_{\mathrm{train}}$ &  &  &  &  &  &  &  &  &  &  \\
\midrule
\endhead
\midrule
\multicolumn{12}{r}{Continued on next page} \\
\midrule
\endfoot
\bottomrule
\endlastfoot
\multirow[t]{4}{*}{MVN 0} & 1000 & 4.548e-01 & 3.508e-02 & 4.121e-01 & 4.071e-02 & 5.057e-01 & 5.581e-02 & 5.344e-01 & 9.255e-02 & 9.639e-01 & 1.910e-02 \\
 & 2000 & 4.658e-01 & 4.352e-02 & 4.121e-01 & 3.872e-02 & 5.179e-01 & 5.411e-02 & 5.441e-01 & 5.511e-02 & 9.790e-01 & 2.515e-02 \\
 & 3000 & 4.560e-01 & 2.904e-02 & 4.122e-01 & 4.682e-02 & 5.249e-01 & 3.699e-02 & 5.592e-01 & 7.778e-02 & 9.976e-01 & 1.733e-03 \\
 & 4000 & 4.883e-01 & 5.100e-02 & 4.116e-01 & 4.091e-02 & 5.295e-01 & 4.251e-02 & 5.595e-01 & 7.586e-02 & 9.975e-01 & 0.000e+00 \\
\cline{1-12}
\multirow[t]{4}{*}{MVN 1} & 1000 & 5.203e-01 & 4.563e-02 & 5.226e-01 & 4.434e-02 & 4.704e-01 & 6.147e-02 & 5.096e-01 & 6.267e-02 & 9.594e-01 & 2.092e-02 \\
 & 2000 & 5.108e-01 & 2.726e-02 & 5.184e-01 & 5.335e-02 & 4.658e-01 & 6.090e-02 & 5.025e-01 & 6.883e-02 & 9.697e-01 & 1.518e-02 \\
 & 3000 & 5.178e-01 & 4.747e-02 & 5.216e-01 & 4.772e-02 & 4.989e-01 & 4.354e-02 & 5.327e-01 & 6.719e-02 & 9.753e-01 & 1.896e-02 \\
 & 4000 & 5.200e-01 & 3.107e-02 & 5.003e-01 & 5.355e-02 & 4.793e-01 & 3.654e-02 & 5.360e-01 & 5.847e-02 & 9.773e-01 & 7.434e-03 \\
\cline{1-12}
\multirow[t]{4}{*}{MVN 2} & 1000 & 4.768e-01 & 5.439e-02 & 4.049e-01 & 3.134e-02 & 5.429e-01 & 5.289e-02 & 5.431e-01 & 6.307e-02 & 9.460e-01 & 5.934e-02 \\
 & 2000 & 4.763e-01 & 4.671e-02 & 4.069e-01 & 3.329e-02 & 5.032e-01 & 6.248e-02 & 5.410e-01 & 7.662e-02 & 9.822e-01 & 5.646e-03 \\
 & 3000 & 4.899e-01 & 6.302e-02 & 4.041e-01 & 3.894e-02 & 5.236e-01 & 7.065e-02 & 5.398e-01 & 6.247e-02 & 9.861e-01 & 8.534e-03 \\
 & 4000 & 5.127e-01 & 7.981e-02 & 4.029e-01 & 3.976e-02 & 5.364e-01 & 5.616e-02 & 5.097e-01 & 7.822e-02 & 9.934e-01 & 2.715e-03 \\
\cline{1-12}
\multirow[t]{4}{*}{SCM 0} & 1000 & 7.471e-01 & 3.716e-02 & 6.218e-01 & 7.268e-02 & 5.805e-01 & 8.022e-02 & 4.739e-01 & 6.207e-02 & 6.113e-01 & 2.409e-02 \\
 & 2000 & 7.250e-01 & 3.665e-02 & 6.024e-01 & 8.207e-02 & 6.618e-01 & 5.272e-02 & 5.038e-01 & 6.840e-02 & 5.945e-01 & 1.699e-02 \\
 & 3000 & 7.387e-01 & 3.545e-02 & 5.639e-01 & 7.701e-02 & 6.381e-01 & 6.953e-02 & 5.201e-01 & 7.332e-02 & 5.973e-01 & 8.906e-03 \\
 & 4000 & 7.071e-01 & 5.140e-02 & 6.099e-01 & 6.310e-02 & 6.683e-01 & 3.808e-02 & 5.694e-01 & 6.003e-02 & 5.996e-01 & 1.497e-02 \\
\cline{1-12}
\multirow[t]{4}{*}{SCM 1} & 1000 & 3.775e-01 & 8.564e-02 & 3.476e-01 & 4.642e-02 & 5.406e-01 & 1.419e-01 & 5.655e-01 & 7.674e-02 & 8.255e-01 & 2.569e-02 \\
 & 2000 & 3.938e-01 & 7.435e-02 & 3.567e-01 & 5.796e-02 & 5.463e-01 & 1.173e-01 & 5.564e-01 & 7.292e-02 & 8.420e-01 & 2.206e-02 \\
 & 3000 & 3.908e-01 & 5.154e-02 & 3.409e-01 & 3.738e-02 & 4.957e-01 & 8.539e-02 & 5.550e-01 & 5.998e-02 & 8.288e-01 & 1.437e-02 \\
 & 4000 & 3.671e-01 & 4.565e-02 & 3.280e-01 & 6.537e-02 & 4.369e-01 & 5.861e-02 & 5.946e-01 & 8.459e-02 & 7.921e-01 & 1.123e-02 \\
\cline{1-12}
\multirow[t]{4}{*}{SCM 2} & 1000 & 4.575e-01 & 5.500e-02 & 4.058e-01 & 6.178e-02 & 5.050e-01 & 8.119e-02 & 5.489e-01 & 6.800e-02 & 6.656e-01 & 2.108e-02 \\
 & 2000 & 4.364e-01 & 4.861e-02 & 4.013e-01 & 5.555e-02 & 4.706e-01 & 6.578e-02 & 5.433e-01 & 6.028e-02 & 6.486e-01 & 1.562e-02 \\
 & 3000 & 4.714e-01 & 4.357e-02 & 4.139e-01 & 6.014e-02 & 4.938e-01 & 8.293e-02 & 5.456e-01 & 6.480e-02 & 6.692e-01 & 1.296e-02 \\
 & 4000 & 4.762e-01 & 4.352e-02 & 4.077e-01 & 7.953e-02 & 5.308e-01 & 5.504e-02 & 5.507e-01 & 5.860e-02 & 6.641e-01 & 1.610e-02 \\
\cline{1-12}
\end{longtable}

\end{landscape}

\begin{longtable}{lllrrrrr}
\caption{
    Predictive performance in the form of $R^2$, for six different datasets and five different models.
    We compare the performance when the graph is fully connected versus when it is pruned to its true edges only.
    The statistics $\mu$ and $\sigma$ represent the mean and standard deviation aggregated over seeds and cross validations.
    With the Linear Mixed Effects (LME) Model we test if the $R^2$ results obtained with the pruned graph are greater than those with the fully connected graph.
    See \cref{fig:results_r2score_all} for the same results.
} \label{tab:results_r2score} \\
\toprule
 &  & Graph & \multicolumn{2}{l}{Fully connected} & \multicolumn{2}{l}{True edges only} & LME \\
 &  & Statistic & $\mu$ & $\sigma$ & $\mu$ & $\sigma$ & $p$-value \\
Dataset & Model & $n_{\mathrm{train}}$ &  &  &  &  &  \\
\midrule
\endfirsthead
\caption[]{
    Predictive performance in the form of $R^2$, for six different datasets and five different models.
    We compare the performance when the graph is fully connected versus when it is pruned to its true edges only.
    The statistics $\mu$ and $\sigma$ represent the mean and standard deviation aggregated over seeds and cross validations.
    With the Linear Mixed Effects (LME) Model we test if the $R^2$ results obtained with the pruned graph are greater than those with the fully connected graph.
    See \cref{fig:results_r2score_all} for the same results.
} \\
\toprule
 &  & Graph & \multicolumn{2}{l}{Fully connected} & \multicolumn{2}{l}{True edges only} & LME \\
 &  & Statistic & $\mu$ & $\sigma$ & $\mu$ & $\sigma$ & $p$-value \\
Dataset & Model & $n_{\mathrm{train}}$ &  &  &  &  &  \\
\midrule
\endhead
\midrule
\multicolumn{8}{r}{Continued on next page} \\
\midrule
\endfoot
\bottomrule
\endlastfoot
\multirow[t]{28}{*}{MVN 0} & \multirow[t]{4}{*}{FTTransformer} & 1000 & 7.928e-01 & 2.000e-03 & 7.950e-01 & 1.782e-03 & 1.221e-07 \\
 &  & 2000 & 7.910e-01 & 7.379e-03 & 7.935e-01 & 3.681e-03 & 9.239e-03 \\
 &  & 3000 & 7.957e-01 & 1.114e-03 & 7.965e-01 & 1.374e-03 & 1.568e-02 \\
 &  & 4000 & 7.975e-01 & 7.178e-04 & 7.983e-01 & 1.299e-03 & 5.861e-02 \\
\cline{2-8}
 & \multirow[t]{4}{*}{FiGNN} & 1000 & 7.883e-01 & 3.955e-03 & 7.879e-01 & 3.121e-03 & 6.958e-01 \\
 &  & 2000 & 7.922e-01 & 2.499e-03 & 7.908e-01 & 2.436e-03 & 9.934e-01 \\
 &  & 3000 & 7.948e-01 & 2.658e-03 & 7.938e-01 & 1.819e-03 & 9.122e-01 \\
 &  & 4000 & 7.952e-01 & 1.693e-03 & 7.951e-01 & 1.575e-03 & 5.715e-01 \\
\cline{2-8}
 & \multirow[t]{4}{*}{INCE} & 1000 & 7.830e-01 & 4.252e-03 & 7.857e-01 & 9.369e-03 & 1.698e-02 \\
 &  & 2000 & 7.863e-01 & 5.282e-03 & 7.916e-01 & 3.636e-03 & 1.403e-07 \\
 &  & 3000 & 7.911e-01 & 2.362e-03 & 7.936e-01 & 1.393e-03 & 1.593e-05 \\
 &  & 4000 & 7.918e-01 & 1.664e-03 & 7.929e-01 & 2.931e-03 & 1.519e-01 \\
\cline{2-8}
 & \multirow[t]{4}{*}{T2GFormer} & 1000 & 7.611e-01 & 2.067e-02 & 7.810e-01 & 1.160e-02 & 3.768e-08 \\
 &  & 2000 & 7.741e-01 & 1.252e-02 & 7.888e-01 & 5.172e-03 & 1.383e-09 \\
 &  & 3000 & 7.830e-01 & 9.613e-03 & 7.894e-01 & 4.455e-03 & 9.891e-04 \\
 &  & 4000 & 7.851e-01 & 8.265e-03 & 7.897e-01 & 5.989e-03 & 8.067e-02 \\
\cline{2-8}
 & \multirow[t]{4}{*}{TabPFN} & 1000 & 7.970e-01 & 1.230e-03 &  &  &  \\
 &  & 2000 & 7.985e-01 & 2.419e-04 &  &  &  \\
 &  & 3000 & 7.991e-01 & 1.497e-04 &  &  &  \\
 &  & 4000 & 7.989e-01 & 6.862e-05 &  &  &  \\
\cline{2-8}
 & \multirow[t]{4}{*}{XGBoost} & 1000 & 7.772e-01 & 7.654e-03 &  &  &  \\
 &  & 2000 & 7.867e-01 & 1.211e-03 &  &  &  \\
 &  & 3000 & 7.896e-01 & 9.442e-04 &  &  &  \\
 &  & 4000 & 7.843e-01 & 1.433e-03 &  &  &  \\
\cline{2-8}
 & \multirow[t]{4}{*}{BDgraph} & 1000 & 7.990e-01 & 2.730e-04 &  &  &  \\
 &  & 2000 & 7.994e-01 & 9.546e-05 &  &  &  \\
 &  & 3000 & 7.993e-01 & 3.010e-04 &  &  &  \\
 &  & 4000 & 7.994e-01 & 6.052e-06 &  &  &  \\
\cline{1-8} \cline{2-8}
\multirow[t]{28}{*}{MVN 1} & \multirow[t]{4}{*}{FTTransformer} & 1000 & 2.873e-01 & 1.157e-02 & 3.008e-01 & 7.322e-03 & 1.013e-12 \\
 &  & 2000 & 3.001e-01 & 4.982e-03 & 3.045e-01 & 4.561e-03 & 9.286e-05 \\
 &  & 3000 & 3.008e-01 & 6.024e-03 & 3.097e-01 & 3.192e-03 & 6.228e-10 \\
 &  & 4000 & 3.072e-01 & 3.170e-03 & 3.059e-01 & 3.257e-03 & 8.191e-01 \\
\cline{2-8}
 & \multirow[t]{4}{*}{FiGNN} & 1000 & 2.914e-01 & 9.587e-03 & 2.884e-01 & 1.186e-02 & 8.939e-01 \\
 &  & 2000 & 2.932e-01 & 1.712e-02 & 2.984e-01 & 6.712e-03 & 4.367e-02 \\
 &  & 3000 & 3.031e-01 & 7.763e-03 & 2.973e-01 & 2.127e-02 & 8.810e-01 \\
 &  & 4000 & 3.053e-01 & 3.281e-03 & 3.071e-01 & 2.695e-03 & 8.243e-02 \\
\cline{2-8}
 & \multirow[t]{4}{*}{INCE} & 1000 & 2.688e-01 & 1.387e-02 & 2.953e-01 & 6.880e-03 & 1.989e-35 \\
 &  & 2000 & 2.823e-01 & 6.577e-03 & 3.009e-01 & 5.433e-03 & 1.105e-43 \\
 &  & 3000 & 2.952e-01 & 7.278e-03 & 3.073e-01 & 5.125e-03 & 2.980e-17 \\
 &  & 4000 & 2.991e-01 & 3.093e-03 & 3.080e-01 & 1.606e-03 & 2.957e-16 \\
\cline{2-8}
 & \multirow[t]{4}{*}{T2GFormer} & 1000 & 2.626e-01 & 2.122e-02 & 3.002e-01 & 9.836e-03 & 2.980e-26 \\
 &  & 2000 & 2.767e-01 & 1.393e-02 & 2.959e-01 & 1.383e-02 & 1.249e-09 \\
 &  & 3000 & 2.927e-01 & 1.595e-02 & 3.082e-01 & 3.218e-03 & 1.576e-06 \\
 &  & 4000 & 3.023e-01 & 7.366e-03 & 3.090e-01 & 5.987e-03 & 1.329e-02 \\
\cline{2-8}
 & \multirow[t]{4}{*}{TabPFN} & 1000 & 3.103e-01 & 2.905e-03 &  &  &  \\
 &  & 2000 & 3.146e-01 & 1.909e-03 &  &  &  \\
 &  & 3000 & 3.151e-01 & 3.399e-04 &  &  &  \\
 &  & 4000 & 3.149e-01 & 2.604e-04 &  &  &  \\
\cline{2-8}
 & \multirow[t]{4}{*}{XGBoost} & 1000 & 2.788e-01 & 8.729e-03 &  &  &  \\
 &  & 2000 & 2.894e-01 & 3.750e-03 &  &  &  \\
 &  & 3000 & 2.996e-01 & 4.644e-03 &  &  &  \\
 &  & 4000 & 2.949e-01 & 3.129e-03 &  &  &  \\
\cline{2-8}
 & \multirow[t]{4}{*}{BDgraph} & 1000 & 3.145e-01 & 1.158e-03 &  &  &  \\
 &  & 2000 & 3.150e-01 & 1.218e-03 &  &  &  \\
 &  & 3000 & 3.151e-01 & 8.791e-04 &  &  &  \\
 &  & 4000 & 3.155e-01 & 2.023e-05 &  &  &  \\
\cline{1-8} \cline{2-8}
\multirow[t]{28}{*}{MVN 2} & \multirow[t]{4}{*}{FTTransformer} & 1000 & 3.111e-01 & 1.771e-02 & 3.218e-01 & 1.356e-02 & 4.767e-06 \\
 &  & 2000 & 3.265e-01 & 5.374e-03 & 3.196e-01 & 1.063e-02 & 9.999e-01 \\
 &  & 3000 & 3.271e-01 & 2.848e-03 & 3.273e-01 & 3.762e-03 & 4.482e-01 \\
 &  & 4000 & 2.952e-01 & 1.051e-01 & 3.331e-01 & 2.340e-03 & 1.272e-01 \\
\cline{2-8}
 & \multirow[t]{4}{*}{FiGNN} & 1000 & 3.110e-01 & 8.880e-03 & 3.069e-01 & 1.322e-02 & 9.635e-01 \\
 &  & 2000 & 3.207e-01 & 5.036e-03 & 3.198e-01 & 4.937e-03 & 7.556e-01 \\
 &  & 3000 & 3.280e-01 & 4.337e-03 & 3.240e-01 & 3.670e-03 & 9.993e-01 \\
 &  & 4000 & 3.290e-01 & 2.607e-03 & 3.273e-01 & 2.265e-03 & 9.404e-01 \\
\cline{2-8}
 & \multirow[t]{4}{*}{INCE} & 1000 & 2.751e-01 & 2.681e-02 & 2.985e-01 & 2.614e-02 & 1.103e-08 \\
 &  & 2000 & 2.977e-01 & 1.265e-02 & 3.040e-01 & 1.372e-02 & 8.489e-03 \\
 &  & 3000 & 3.159e-01 & 5.979e-03 & 3.211e-01 & 5.549e-03 & 8.908e-04 \\
 &  & 4000 & 3.177e-01 & 6.735e-03 & 3.283e-01 & 2.787e-03 & 2.291e-06 \\
\cline{2-8}
 & \multirow[t]{4}{*}{T2GFormer} & 1000 & 2.725e-01 & 3.057e-02 & 2.964e-01 & 1.609e-02 & 9.802e-07 \\
 &  & 2000 & 2.904e-01 & 2.975e-02 & 3.150e-01 & 1.239e-02 & 3.412e-08 \\
 &  & 3000 & 3.140e-01 & 1.346e-02 & 3.203e-01 & 7.734e-03 & 1.832e-02 \\
 &  & 4000 & 3.049e-01 & 1.888e-02 & 3.193e-01 & 1.188e-02 & 2.071e-02 \\
\cline{2-8}
 & \multirow[t]{4}{*}{TabPFN} & 1000 & 3.035e-01 & 1.660e-03 &  &  &  \\
 &  & 2000 & 3.046e-01 & 5.855e-04 &  &  &  \\
 &  & 3000 & 3.048e-01 & 3.652e-04 &  &  &  \\
 &  & 4000 & 3.061e-01 & 3.168e-04 &  &  &  \\
\cline{2-8}
 & \multirow[t]{4}{*}{XGBoost} & 1000 & 2.641e-01 & 2.197e-02 &  &  &  \\
 &  & 2000 & 2.789e-01 & 3.089e-03 &  &  &  \\
 &  & 3000 & 2.890e-01 & 3.739e-03 &  &  &  \\
 &  & 4000 & 2.889e-01 & 2.763e-03 &  &  &  \\
\cline{2-8}
 & \multirow[t]{4}{*}{BDgraph} & 1000 & 3.376e-01 & 1.335e-03 &  &  &  \\
 &  & 2000 & 3.392e-01 & 2.941e-04 &  &  &  \\
 &  & 3000 & 3.380e-01 & 1.755e-04 &  &  &  \\
 &  & 4000 & 3.380e-01 & 1.424e-05 &  &  &  \\
\cline{1-8} \cline{2-8}
\multirow[t]{28}{*}{SCM 0} & \multirow[t]{4}{*}{FTTransformer} & 1000 & 6.852e-01 & 1.038e-02 & 7.012e-01 & 8.936e-03 & 3.191e-14 \\
 &  & 2000 & 6.996e-01 & 1.954e-02 & 7.195e-01 & 7.664e-03 & 5.864e-12 \\
 &  & 3000 & 7.166e-01 & 5.022e-03 & 7.186e-01 & 3.711e-03 & 6.297e-02 \\
 &  & 4000 & 7.176e-01 & 4.316e-03 & 7.191e-01 & 3.196e-03 & 1.897e-01 \\
\cline{2-8}
 & \multirow[t]{4}{*}{FiGNN} & 1000 & 6.857e-01 & 8.771e-03 & 6.770e-01 & 1.356e-02 & 9.999e-01 \\
 &  & 2000 & 6.987e-01 & 5.186e-03 & 6.896e-01 & 1.378e-02 & 9.998e-01 \\
 &  & 3000 & 6.985e-01 & 1.175e-02 & 7.051e-01 & 4.950e-03 & 7.274e-03 \\
 &  & 4000 & 7.122e-01 & 3.105e-03 & 7.129e-01 & 4.739e-03 & 3.559e-01 \\
\cline{2-8}
 & \multirow[t]{4}{*}{INCE} & 1000 & 6.169e-01 & 1.583e-02 & 6.934e-01 & 1.276e-02 & 7.188e-139 \\
 &  & 2000 & 6.769e-01 & 1.258e-02 & 7.155e-01 & 3.774e-03 & 8.700e-70 \\
 &  & 3000 & 6.881e-01 & 8.860e-03 & 7.149e-01 & 5.249e-03 & 4.991e-32 \\
 &  & 4000 & 6.993e-01 & 1.378e-02 & 7.200e-01 & 3.107e-03 & 1.913e-06 \\
\cline{2-8}
 & \multirow[t]{4}{*}{T2GFormer} & 1000 & 6.064e-01 & 6.247e-02 & 6.952e-01 & 1.588e-02 & 1.506e-18 \\
 &  & 2000 & 6.792e-01 & 1.816e-02 & 7.134e-01 & 9.392e-03 & 4.987e-22 \\
 &  & 3000 & 7.027e-01 & 9.075e-03 & 7.054e-01 & 1.035e-02 & 1.934e-01 \\
 &  & 4000 & 7.079e-01 & 5.211e-03 & 7.218e-01 & 3.525e-03 & 1.184e-12 \\
\cline{2-8}
 & \multirow[t]{4}{*}{TabPFN} & 1000 & 7.199e-01 & 3.512e-03 &  &  &  \\
 &  & 2000 & 7.260e-01 & 1.418e-03 &  &  &  \\
 &  & 3000 & 7.278e-01 & 4.063e-04 &  &  &  \\
 &  & 4000 & 7.288e-01 & 1.264e-04 &  &  &  \\
\cline{2-8}
 & \multirow[t]{4}{*}{XGBoost} & 1000 & 6.038e-01 & 2.008e-02 &  &  &  \\
 &  & 2000 & 6.360e-01 & 4.754e-03 &  &  &  \\
 &  & 3000 & 6.582e-01 & 1.142e-02 &  &  &  \\
 &  & 4000 & 6.790e-01 & 3.240e-03 &  &  &  \\
\cline{2-8}
 & \multirow[t]{4}{*}{BDgraph} & 1000 & 2.865e-01 & 4.755e-03 &  &  &  \\
 &  & 2000 & 2.895e-01 & 4.179e-03 &  &  &  \\
 &  & 3000 & 2.903e-01 & 1.251e-03 &  &  &  \\
 &  & 4000 & 2.944e-01 & 9.584e-05 &  &  &  \\
\cline{1-8} \cline{2-8}
\multirow[t]{28}{*}{SCM 1} & \multirow[t]{4}{*}{FTTransformer} & 1000 & 8.421e-01 & 4.800e-03 & 8.476e-01 & 6.136e-03 & 3.358e-07 \\
 &  & 2000 & 8.530e-01 & 3.025e-03 & 8.578e-01 & 3.103e-03 & 1.220e-12 \\
 &  & 3000 & 8.536e-01 & 8.851e-03 & 8.614e-01 & 3.758e-03 & 2.128e-07 \\
 &  & 4000 & 8.655e-01 & 1.603e-03 & 8.629e-01 & 3.397e-03 & 9.864e-01 \\
\cline{2-8}
 & \multirow[t]{4}{*}{FiGNN} & 1000 & 8.353e-01 & 1.802e-02 & 8.456e-01 & 4.649e-03 & 6.706e-05 \\
 &  & 2000 & 8.544e-01 & 3.347e-03 & 8.549e-01 & 2.867e-03 & 2.557e-01 \\
 &  & 3000 & 8.575e-01 & 4.599e-03 & 8.579e-01 & 5.205e-03 & 3.690e-01 \\
 &  & 4000 & 8.601e-01 & 1.891e-03 & 8.618e-01 & 2.076e-03 & 3.270e-02 \\
\cline{2-8}
 & \multirow[t]{4}{*}{INCE} & 1000 & 8.407e-01 & 4.241e-03 & 8.483e-01 & 4.130e-03 & 1.477e-19 \\
 &  & 2000 & 8.489e-01 & 3.068e-03 & 8.532e-01 & 3.350e-03 & 5.220e-08 \\
 &  & 3000 & 8.549e-01 & 2.966e-03 & 8.593e-01 & 2.800e-03 & 7.280e-08 \\
 &  & 4000 & 8.572e-01 & 3.749e-03 & 8.629e-01 & 1.749e-03 & 6.160e-06 \\
\cline{2-8}
 & \multirow[t]{4}{*}{T2GFormer} & 1000 & 8.302e-01 & 6.270e-03 & 8.286e-01 & 1.756e-02 & 7.166e-01 \\
 &  & 2000 & 8.402e-01 & 6.429e-03 & 8.444e-01 & 7.657e-03 & 9.535e-03 \\
 &  & 3000 & 8.475e-01 & 5.003e-03 & 8.444e-01 & 1.111e-02 & 8.923e-01 \\
 &  & 4000 & 8.508e-01 & 5.679e-03 & 8.540e-01 & 3.901e-03 & 7.649e-02 \\
\cline{2-8}
 & \multirow[t]{4}{*}{TabPFN} & 1000 & 8.721e-01 & 1.974e-03 &  &  &  \\
 &  & 2000 & 8.756e-01 & 9.270e-04 &  &  &  \\
 &  & 3000 & 8.780e-01 & 3.718e-04 &  &  &  \\
 &  & 4000 & 8.783e-01 & 2.596e-04 &  &  &  \\
\cline{2-8}
 & \multirow[t]{4}{*}{XGBoost} & 1000 & 8.495e-01 & 4.047e-03 &  &  &  \\
 &  & 2000 & 8.598e-01 & 1.057e-03 &  &  &  \\
 &  & 3000 & 8.618e-01 & 7.716e-04 &  &  &  \\
 &  & 4000 & 8.642e-01 & 7.016e-04 &  &  &  \\
\cline{2-8}
 & \multirow[t]{4}{*}{BDgraph} & 1000 & 7.968e-01 & 1.242e-03 &  &  &  \\
 &  & 2000 & 7.973e-01 & 5.478e-04 &  &  &  \\
 &  & 3000 & 7.972e-01 & 1.416e-04 &  &  &  \\
 &  & 4000 & 7.974e-01 & 1.145e-05 &  &  &  \\
\cline{1-8} \cline{2-8}
\multirow[t]{28}{*}{SCM 2} & \multirow[t]{4}{*}{FTTransformer} & 1000 & 7.988e-01 & 6.280e-03 & 8.068e-01 & 5.799e-03 & 2.428e-11 \\
 &  & 2000 & 8.069e-01 & 5.623e-03 & 8.118e-01 & 3.220e-03 & 2.246e-07 \\
 &  & 3000 & 8.122e-01 & 2.971e-03 & 8.158e-01 & 1.102e-03 & 2.046e-08 \\
 &  & 4000 & 8.137e-01 & 9.271e-04 & 8.171e-01 & 1.481e-03 & 4.620e-10 \\
\cline{2-8}
 & \multirow[t]{4}{*}{FiGNN} & 1000 & 7.913e-01 & 1.540e-02 & 7.994e-01 & 1.198e-02 & 1.783e-04 \\
 &  & 2000 & 8.093e-01 & 5.427e-03 & 8.112e-01 & 6.110e-03 & 5.826e-02 \\
 &  & 3000 & 8.152e-01 & 3.812e-03 & 8.073e-01 & 1.318e-02 & 9.948e-01 \\
 &  & 4000 & 8.162e-01 & 1.393e-03 & 8.166e-01 & 1.864e-03 & 2.944e-01 \\
\cline{2-8}
 & \multirow[t]{4}{*}{INCE} & 1000 & 7.922e-01 & 6.950e-03 & 8.021e-01 & 5.126e-03 & 7.559e-20 \\
 &  & 2000 & 8.033e-01 & 3.749e-03 & 8.054e-01 & 5.855e-03 & 4.815e-02 \\
 &  & 3000 & 8.068e-01 & 4.149e-03 & 8.137e-01 & 3.154e-03 & 2.383e-18 \\
 &  & 4000 & 8.069e-01 & 3.399e-03 & 8.132e-01 & 2.133e-03 & 2.872e-07 \\
\cline{2-8}
 & \multirow[t]{4}{*}{T2GFormer} & 1000 & 7.863e-01 & 1.251e-02 & 7.923e-01 & 1.657e-02 & 2.163e-02 \\
 &  & 2000 & 7.974e-01 & 1.011e-02 & 8.084e-01 & 4.063e-03 & 1.483e-09 \\
 &  & 3000 & 7.993e-01 & 9.041e-03 & 8.115e-01 & 4.340e-03 & 2.233e-09 \\
 &  & 4000 & 8.020e-01 & 8.168e-03 & 8.125e-01 & 4.053e-03 & 1.372e-04 \\
\cline{2-8}
 & \multirow[t]{4}{*}{TabPFN} & 1000 & 7.972e-01 & 1.963e-03 &  &  &  \\
 &  & 2000 & 8.024e-01 & 1.394e-03 &  &  &  \\
 &  & 3000 & 8.035e-01 & 1.564e-03 &  &  &  \\
 &  & 4000 & 8.042e-01 & 1.431e-03 &  &  &  \\
\cline{2-8}
 & \multirow[t]{4}{*}{XGBoost} & 1000 & 7.725e-01 & 5.500e-03 &  &  &  \\
 &  & 2000 & 7.858e-01 & 1.684e-03 &  &  &  \\
 &  & 3000 & 7.908e-01 & 1.714e-03 &  &  &  \\
 &  & 4000 & 7.896e-01 & 1.285e-03 &  &  &  \\
\cline{2-8}
 & \multirow[t]{4}{*}{BDgraph} & 1000 & 7.575e-01 & 1.307e-03 &  &  &  \\
 &  & 2000 & 7.587e-01 & 3.103e-04 &  &  &  \\
 &  & 3000 & 7.606e-01 & 4.838e-05 &  &  &  \\
 &  & 4000 & 7.607e-01 & 1.319e-05 &  &  &  \\
\cline{1-8} \cline{2-8}
\end{longtable}

\FloatBarrier

\subsection{Higher number of training samples} \label{sec:results_ntrain}

\begin{figure}[htbp]
    \centering
    \includegraphics[width=\textwidth]{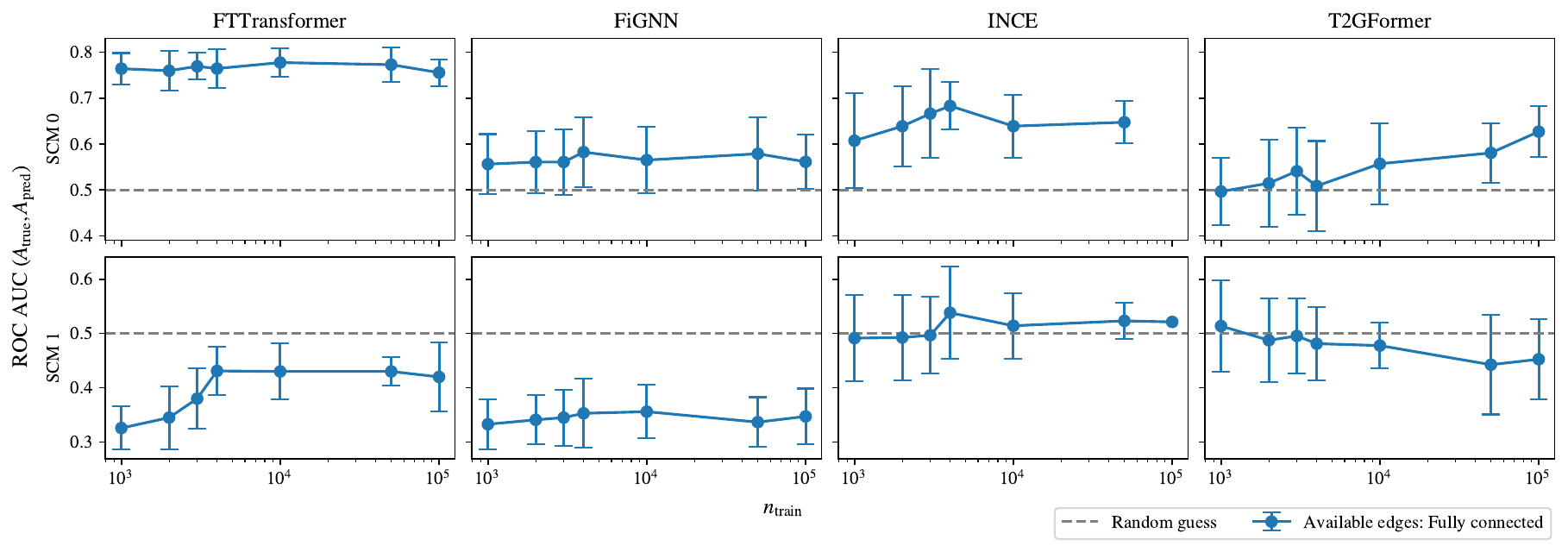}
    \caption{Graph quality in the form of the \roc with a higher number of training samples \ntrain than the main experiment, for two different datasets.
    Error bars show standard deviation across seeds and cross validations.
    }
    \label{fig:results_roc_auc_ntrain}
\end{figure}

\begin{figure}[htbp]
    \centering
    \includegraphics[width=\textwidth]{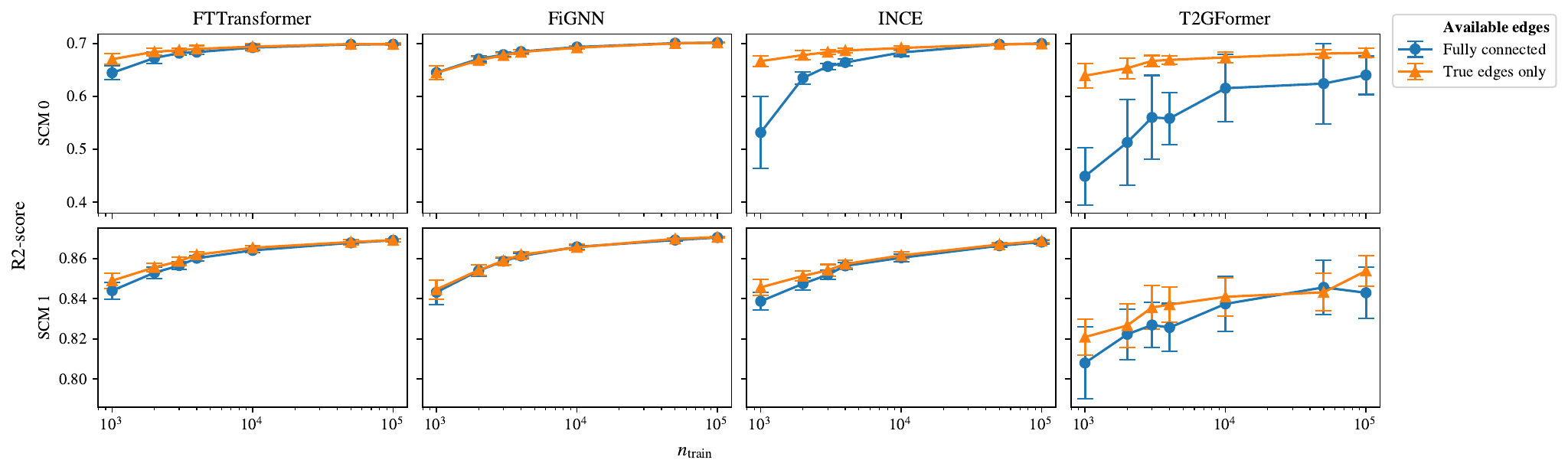}
    \caption{Predictive performance with a higher number of training samples \ntrain than the main experiment, for two different datasets.
    Error bars show standard deviation across seeds and cross validations.
    }
    \label{fig:results_r2_ntrain}
\end{figure}

We repeat the experiments from \cref{sec:analysis} while varying the number of training samples \ntrain up to $10^5$ samples. 
Note that $10^5$ samples is quite high, considering the low number of features $p=10$ and the relatively simple feature interactions in the datasets.
To reduce computational cost, we do not tune the hyperparameters but use the default hyperparameters as previously described in \cref{sec:exp_details}, and we only evaluate two of the six datasets. We do use the same cross-validation strategy with 10 seeds as before. The results for the learned graph structure are shown in \cref{fig:results_roc_auc_ntrain}, results for the predictive performance are shown in \cref{fig:results_r2_ntrain}. The results are consistent with the main experiment (tuned hyperparameters, up to $\ntrain = 4000$ samples), shown in \cref{fig:results_roc_auc_all} and \cref{fig:results_r2score_all}. However, the results have a larger variety because the hyperparameters are not tuned.

\FloatBarrier

\subsection{Node-level versus graph-level models} \label{sec:node-level_graph-level}
\begin{figure}[htbp]
    \centering
    \includegraphics[width=\textwidth]{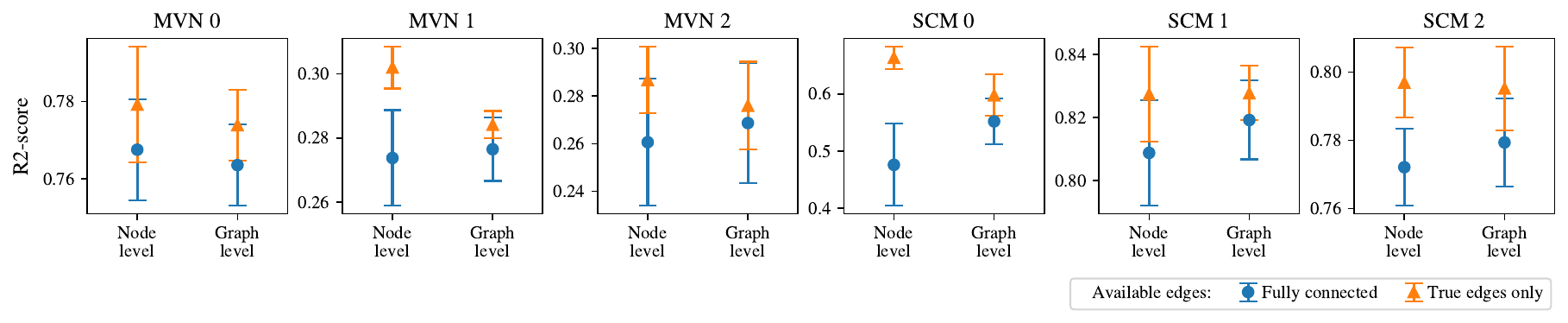}
    \caption{Node-level (default) versus graph-level adaptions of \tg for multiple \mvn and \scm datasets. The node-level adaptation benefits more from pruning the graph to the true edges.
    Error bars show standard deviation across seeds and cross validations.
    }
    \label{fig:node_graph_level}
\end{figure}

In \cref{fig:results_r2score}, not all models benefit from pruning the graph. For \fignn, the pruned and fully connected graphs have similar performance. This could be because \fignn treats the task of predicting the target feature on a graph-level task, while the other models have a target token and treat it as a node-level task. As an example, we adapt the architecture from \tg, which is by default a node-level model, to a graph-level model. Results on the predictive performance are shown in \cref{fig:node_graph_level}. The graph-level model benefits less from pruning the graph than the node-level model. This indicates that the node-level models are more sensitive to the graph structure than the graph-level models.

\FloatBarrier


\begin{thebibliography}{73}
\providecommand{\natexlab}[1]{#1}
\providecommand{\url}[1]{\texttt{#1}}
\expandafter\ifx\csname urlstyle\endcsname\relax
  \providecommand{\doi}[1]{doi: #1}\else
  \providecommand{\doi}{doi: \begingroup \urlstyle{rm}\Url}\fi

\bibitem[Akiba et~al.(2019)Akiba, Sano, Yanase, Ohta, and Koyama]{akibaOptunaNextgenerationHyperparameter2019}
Takuya Akiba, Shotaro Sano, Toshihiko Yanase, Takeru Ohta, and Masanori Koyama.
\newblock Optuna: {{A Next-generation Hyperparameter Optimization Framework}}.
\newblock In \emph{Proceedings of the 25th {{ACM SIGKDD International Conference}} on {{Knowledge Discovery}} \& {{Data Mining}}}, pp.\  2623--2631, Anchorage AK USA, July 2019. ACM.
\newblock ISBN 978-1-4503-6201-6.
\newblock \doi{10.1145/3292500.3330701}.
\newblock URL \url{https://dl.acm.org/doi/10.1145/3292500.3330701}.

\bibitem[Arik \& Pfister(2020)Arik and Pfister]{arikTabNetAttentiveInterpretable2020}
Sercan~O. Arik and Tomas Pfister.
\newblock {{TabNet}}: {{Attentive Interpretable Tabular Learning}}, December 2020.
\newblock URL \url{http://arxiv.org/abs/1908.07442}.

\bibitem[Battaglia et~al.(2016)Battaglia, Pascanu, Lai, Rezende, and Kavukcuoglu]{battagliaInteractionNetworksLearning2016}
Peter~W. Battaglia, Razvan Pascanu, Matthew Lai, Danilo Rezende, and Koray Kavukcuoglu.
\newblock Interaction {{Networks}} for {{Learning}} about {{Objects}}, {{Relations}} and {{Physics}}, December 2016.
\newblock URL \url{http://arxiv.org/abs/1612.00222}.

\bibitem[Battaglia et~al.(2018)Battaglia, Hamrick, Bapst, {Sanchez-Gonzalez}, Zambaldi, Malinowski, Tacchetti, Raposo, Santoro, Faulkner, Gulcehre, Song, Ballard, Gilmer, Dahl, Vaswani, Allen, Nash, Langston, Dyer, Heess, Wierstra, Kohli, Botvinick, Vinyals, Li, and Pascanu]{battagliaRelationalInductiveBiases2018}
Peter~W. Battaglia, Jessica~B. Hamrick, Victor Bapst, Alvaro {Sanchez-Gonzalez}, Vinicius Zambaldi, Mateusz Malinowski, Andrea Tacchetti, David Raposo, Adam Santoro, Ryan Faulkner, Caglar Gulcehre, Francis Song, Andrew Ballard, Justin Gilmer, George Dahl, Ashish Vaswani, Kelsey Allen, Charles Nash, Victoria Langston, Chris Dyer, Nicolas Heess, Daan Wierstra, Pushmeet Kohli, Matt Botvinick, Oriol Vinyals, Yujia Li, and Razvan Pascanu.
\newblock Relational inductive biases, deep learning, and graph networks, October 2018.
\newblock URL \url{http://arxiv.org/abs/1806.01261}.

\bibitem[Bergstra et~al.(2011)Bergstra, Bardenet, Bengio, and K{\'e}gl]{bergstraAlgorithmsHyperParameterOptimization2011}
James Bergstra, R{\'e}mi Bardenet, Yoshua Bengio, and Bal{\'a}zs K{\'e}gl.
\newblock Algorithms for {{Hyper-Parameter Optimization}}.
\newblock In \emph{Advances in {{Neural Information Processing Systems}}}, volume~24. Curran Associates, Inc., 2011.
\newblock URL \url{https://papers.nips.cc/paper_files/paper/2011/hash/86e8f7ab32cfd12577bc2619bc635690-Abstract.html}.

\bibitem[Bibal et~al.(2022)Bibal, Cardon, Alfter, Wilkens, Wang, Fran{\c c}ois, and Watrin]{bibalAttentionExplanationIntroduction2022}
Adrien Bibal, R{\'e}mi Cardon, David Alfter, Rodrigo Wilkens, Xiaoou Wang, Thomas Fran{\c c}ois, and Patrick Watrin.
\newblock Is {{Attention Explanation}}? {{An Introduction}} to the {{Debate}}.
\newblock In Smaranda Muresan, Preslav Nakov, and Aline Villavicencio (eds.), \emph{Proceedings of the 60th {{Annual Meeting}} of the {{Association}} for {{Computational Linguistics}} ({{Volume}} 1: {{Long Papers}})}, pp.\  3889--3900, Dublin, Ireland, May 2022. Association for Computational Linguistics.
\newblock \doi{10.18653/v1/2022.acl-long.269}.
\newblock URL \url{https://aclanthology.org/2022.acl-long.269/}.

\bibitem[Bradley(1997)]{bradleyUseAreaROC1997}
Andrew~P. Bradley.
\newblock The use of the area under the {{ROC}} curve in the evaluation of machine learning algorithms.
\newblock \emph{Pattern Recognition}, 30\penalty0 (7):\penalty0 1145--1159, July 1997.
\newblock ISSN 00313203.
\newblock \doi{10.1016/S0031-3203(96)00142-2}.
\newblock URL \url{https://linkinghub.elsevier.com/retrieve/pii/S0031320396001422}.

\bibitem[Bronstein et~al.(2021)Bronstein, Bruna, Cohen, and Veli{\v c}kovi{\'c}]{bronsteinGeometricDeepLearning2021}
Michael~M. Bronstein, Joan Bruna, Taco Cohen, and Petar Veli{\v c}kovi{\'c}.
\newblock Geometric {{Deep Learning}}: {{Grids}}, {{Groups}}, {{Graphs}}, {{Geodesics}}, and {{Gauges}}, May 2021.
\newblock URL \url{http://arxiv.org/abs/2104.13478}.

\bibitem[Cai et~al.(2021)Cai, Zheng, Chen, Jagadish, Ooi, and Zhang]{caiARMNetAdaptiveRelation2021}
Shaofeng Cai, Kaiping Zheng, Gang Chen, H.~V. Jagadish, Beng~Chin Ooi, and Meihui Zhang.
\newblock {{ARM-Net}}: {{Adaptive Relation Modeling Network}} for {{Structured Data}}.
\newblock In \emph{Proceedings of the 2021 {{International Conference}} on {{Management}} of {{Data}}}, pp.\  207--220, Virtual Event China, June 2021. ACM.
\newblock ISBN 978-1-4503-8343-1.
\newblock \doi{10.1145/3448016.3457321}.
\newblock URL \url{https://dl.acm.org/doi/10.1145/3448016.3457321}.

\bibitem[Chen \& Guestrin(2016)Chen and Guestrin]{chenXGBoostScalableTree2016}
Tianqi Chen and Carlos Guestrin.
\newblock {{XGBoost}}: {{A Scalable Tree Boosting System}}.
\newblock In \emph{Proceedings of the 22nd {{ACM SIGKDD International Conference}} on {{Knowledge Discovery}} and {{Data Mining}}}, pp.\  785--794, August 2016.
\newblock \doi{10.1145/2939672.2939785}.
\newblock URL \url{http://arxiv.org/abs/1603.02754}.

\bibitem[Chen et~al.(2026)Chen, He, Benesty, Khotilovich, Tang, Cho, Chen, Mitchell, Cano, Zhou, Li, Xie, Lin, Geng, Li, Yuan, and Cortes]{chenXgboostExtremeGradient2026}
Tianqi Chen, Tong He, Michael Benesty, Vadim Khotilovich, Yuan Tang, Hyunsu Cho, Kailong Chen, Rory Mitchell, Ignacio Cano, Tianyi Zhou, Mu~Li, Junyuan Xie, Min Lin, Yifeng Geng, Yutian Li, Jiaming Yuan, and David Cortes.
\newblock \emph{Xgboost: {{Extreme}} Gradient Boosting}, 2026.
\newblock URL \url{https://github.com/dmlc/xgboost}.

\bibitem[Cheng et~al.(2016)Cheng, Koc, Harmsen, Shaked, Chandra, Aradhye, Anderson, Corrado, Chai, Ispir, Anil, Haque, Hong, Jain, Liu, and Shah]{chengWideDeepLearning2016}
Heng-Tze Cheng, Levent Koc, Jeremiah Harmsen, Tal Shaked, Tushar Chandra, Hrishi Aradhye, Glen Anderson, Greg Corrado, Wei Chai, Mustafa Ispir, Rohan Anil, Zakaria Haque, Lichan Hong, Vihan Jain, Xiaobing Liu, and Hemal Shah.
\newblock Wide \& {{Deep Learning}} for {{Recommender Systems}}, June 2016.
\newblock URL \url{http://arxiv.org/abs/1606.07792}.

\bibitem[Cong et~al.(2024)Cong, Hulsebos, Sun, Groth, and Jagadish]{congObservatoryCharacterizingEmbeddings2024}
Tianji Cong, Madelon Hulsebos, Zhenjie Sun, Paul Groth, and H.~V. Jagadish.
\newblock Observatory: {{Characterizing Embeddings}} of {{Relational Tables}}, January 2024.
\newblock URL \url{http://arxiv.org/abs/2310.07736}.

\bibitem[Cowell et~al.(1999)Cowell, Lauritzen, David, Spiegelhalter, Nair, Lawless, and Jordan]{cowellProbabilisticNetworksExpert1999}
Robert~G. Cowell, Steffen~L. Lauritzen, A.~Philip David, David~J. Spiegelhalter, V.~Nair, J.~Lawless, and M.~Jordan.
\newblock \emph{Probabilistic Networks and Expert Systems}.
\newblock Springer-Verlag, Berlin, Heidelberg, 1 edition, 1999.
\newblock ISBN 0-387-98767-3.

\bibitem[Devlin et~al.(2019)Devlin, Chang, Lee, and Toutanova]{devlinBERTPretrainingDeep2019}
Jacob Devlin, Ming-Wei Chang, Kenton Lee, and Kristina Toutanova.
\newblock {{BERT}}: {{Pre-training}} of {{Deep Bidirectional Transformers}} for {{Language Understanding}}, May 2019.
\newblock URL \url{http://arxiv.org/abs/1810.04805}.

\bibitem[Dwivedi et~al.(2025)Dwivedi, Jaladi, Shen, L{\'o}pez, Kanatsoulis, Puri, Fey, and Leskovec]{dwivediRelationalGraphTransformer2025}
Vijay~Prakash Dwivedi, Sri Jaladi, Yangyi Shen, Federico L{\'o}pez, Charilaos~I. Kanatsoulis, Rishi Puri, Matthias Fey, and Jure Leskovec.
\newblock Relational {{Graph Transformer}}, May 2025.
\newblock URL \url{http://arxiv.org/abs/2505.10960}.

\bibitem[Feurer et~al.(2022)Feurer, Eggensperger, Falkner, Lindauer, and Hutter]{feurerAutoSklearn20Handsfree2022}
Matthias Feurer, Katharina Eggensperger, Stefan Falkner, Marius Lindauer, and Frank Hutter.
\newblock Auto-{{Sklearn}} 2.0: {{Hands-free AutoML}} via {{Meta-Learning}}.
\newblock \emph{Journal of Machine Learning Research}, 23\penalty0 (261):\penalty0 1--61, 2022.
\newblock ISSN 1533-7928.
\newblock URL \url{http://jmlr.org/papers/v23/21-0992.html}.

\bibitem[Fey et~al.(2023)Fey, Hu, Huang, Lenssen, Ranjan, Robinson, Ying, You, and Leskovec]{feyRelationalDeepLearning2023}
Matthias Fey, Weihua Hu, Kexin Huang, Jan~Eric Lenssen, Rishabh Ranjan, Joshua Robinson, Rex Ying, Jiaxuan You, and Jure Leskovec.
\newblock Relational {{Deep Learning}}: {{Graph Representation Learning}} on {{Relational Databases}}, December 2023.
\newblock URL \url{http://arxiv.org/abs/2312.04615}.

\bibitem[Fukushima(1980)]{fukushimaNeocognitronSelforganizingNeural1980}
Kunihiko Fukushima.
\newblock Neocognitron: {{A}} self-organizing neural network model for a mechanism of pattern recognition unaffected by shift in position.
\newblock \emph{Biological Cybernetics}, 36\penalty0 (4):\penalty0 193--202, April 1980.
\newblock ISSN 0340-1200, 1432-0770.
\newblock \doi{10.1007/BF00344251}.
\newblock URL \url{http://link.springer.com/10.1007/BF00344251}.

\bibitem[Gorishniy et~al.(2021)Gorishniy, Rubachev, Khrulkov, and Babenko]{gorishniyRevisitingDeepLearning2021}
Yury Gorishniy, Ivan Rubachev, Valentin Khrulkov, and Artem Babenko.
\newblock Revisiting {{Deep Learning Models}} for {{Tabular Data}}.
\newblock In \emph{Advances in {{Neural Information Processing Systems}}}, volume~34, pp.\  18932--18943. Curran Associates, Inc., 2021.
\newblock URL \url{https://proceedings.neurips.cc/paper_files/paper/2021/hash/9d86d83f925f2149e9edb0ac3b49229c-Abstract.html}.

\bibitem[Goyal \& Bengio(2022)Goyal and Bengio]{goyalInductiveBiasesDeep2022}
Anirudh Goyal and Yoshua Bengio.
\newblock Inductive {{Biases}} for {{Deep Learning}} of {{Higher-Level Cognition}}, August 2022.
\newblock URL \url{http://arxiv.org/abs/2011.15091}.

\bibitem[Grinsztajn et~al.(2022)Grinsztajn, Oyallon, and Varoquaux]{grinsztajnWhyTreebasedModels2022}
L{\'e}o Grinsztajn, Edouard Oyallon, and Ga{\"e}l Varoquaux.
\newblock Why do tree-based models still outperform deep learning on tabular data?, July 2022.
\newblock URL \url{http://arxiv.org/abs/2207.08815}.

\bibitem[Guo et~al.(2017)Guo, Tang, Ye, Li, and He]{guoDeepFMFactorizationMachineBased2017}
Huifeng Guo, Ruiming Tang, Yunming Ye, Zhenguo Li, and Xiuqiang He.
\newblock {{DeepFM}}: {{A Factorization-Machine}} based {{Neural Network}} for {{CTR Prediction}}, March 2017.
\newblock URL \url{http://arxiv.org/abs/1703.04247}.

\bibitem[Hollmann et~al.(2025)Hollmann, M{\"u}ller, Purucker, Krishnakumar, K{\"o}rfer, Hoo, Schirrmeister, and Hutter]{hollmannAccuratePredictionsSmall2025}
Noah Hollmann, Samuel M{\"u}ller, Lennart Purucker, Arjun Krishnakumar, Max K{\"o}rfer, Shi~Bin Hoo, Robin~Tibor Schirrmeister, and Frank Hutter.
\newblock Accurate predictions on small data with a tabular foundation model.
\newblock \emph{Nature}, 637\penalty0 (8045):\penalty0 319--326, January 2025.
\newblock ISSN 0028-0836, 1476-4687.
\newblock \doi{10.1038/s41586-024-08328-6}.
\newblock URL \url{https://www.nature.com/articles/s41586-024-08328-6}.

\bibitem[Huang et~al.(2020)Huang, Khetan, Cvitkovic, and Karnin]{huangTabTransformerTabularData2020}
Xin Huang, Ashish Khetan, Milan Cvitkovic, and Zohar Karnin.
\newblock {{TabTransformer}}: {{Tabular Data Modeling Using Contextual Embeddings}}, December 2020.
\newblock URL \url{http://arxiv.org/abs/2012.06678}.

\bibitem[Joshi(2025)]{joshiTransformersAreGraph2025}
Chaitanya~K. Joshi.
\newblock Transformers are {{Graph Neural Networks}}, June 2025.
\newblock URL \url{http://arxiv.org/abs/2506.22084}.

\bibitem[Ke et~al.(2017)Ke, Meng, Finley, Wang, Chen, Ma, Ye, and Liu]{keLightGBMHighlyEfficient2017}
Guolin Ke, Qi~Meng, Thomas Finley, Taifeng Wang, Wei Chen, Weidong Ma, Qiwei Ye, and Tie-Yan Liu.
\newblock {{LightGBM}}: {{A Highly Efficient Gradient Boosting Decision Tree}}.
\newblock In \emph{Advances in {{Neural Information Processing Systems}}}, volume~30. Curran Associates, Inc., 2017.
\newblock URL \url{https://proceedings.neurips.cc/paper_files/paper/2017/hash/6449f44a102fde848669bdd9eb6b76fa-Abstract.html}.

\bibitem[Kingma \& Ba(2017)Kingma and Ba]{kingmaAdamMethodStochastic2017}
Diederik~P. Kingma and Jimmy Ba.
\newblock Adam: {{A Method}} for {{Stochastic Optimization}}, January 2017.
\newblock URL \url{http://arxiv.org/abs/1412.6980}.

\bibitem[Koller \& Friedman(2009)Koller and Friedman]{kollerProbabilisticGraphicalModels2009}
Daphne Koller and Nir Friedman.
\newblock \emph{Probabilistic {{Graphical Models}}: {{Principles}} and {{Techniques}}}.
\newblock Adaptive {{Computation}} and {{Machine Learning}}. MIT Press, Cambridge, 2009.
\newblock ISBN 978-0-262-25835-7.

\bibitem[Kossen et~al.(2021)Kossen, Band, Lyle, Gomez, Rainforth, and Gal]{kossenSelfAttentionDatapointsGoing2021}
Jannik Kossen, Neil Band, Clare Lyle, Aidan~N Gomez, Thomas Rainforth, and Yarin Gal.
\newblock Self-{{Attention Between Datapoints}}: {{Going Beyond Individual Input-Output Pairs}} in {{Deep Learning}}.
\newblock In \emph{Advances in {{Neural Information Processing Systems}}}, volume~34, pp.\  28742--28756. Curran Associates, Inc., 2021.
\newblock URL \url{https://proceedings.neurips.cc/paper_files/paper/2021/hash/f1507aba9fc82ffa7cc7373c58f8a613-Abstract.html}.

\bibitem[Lauritzen(1996)]{lauritzenGraphicalModels1996}
Steffen~L. Lauritzen.
\newblock \emph{Graphical Models}.
\newblock Number~17 in Oxford Statistical Science Series. Clarendon Press ; Oxford University Press, Oxford : New York, 1996.
\newblock ISBN 978-0-19-852219-5.

\bibitem[LeCun et~al.(1989)LeCun, Boser, Denker, Henderson, Howard, Hubbard, and Jackel]{lecunBackpropagationAppliedHandwritten1989}
Y.~LeCun, B.~Boser, J.~S. Denker, D.~Henderson, R.~E. Howard, W.~Hubbard, and L.~D. Jackel.
\newblock Backpropagation {{Applied}} to {{Handwritten Zip Code Recognition}}.
\newblock \emph{Neural Computation}, 1\penalty0 (4):\penalty0 541--551, December 1989.
\newblock ISSN 0899-7667.
\newblock \doi{10.1162/neco.1989.1.4.541}.
\newblock URL \url{https://ieeexplore.ieee.org/abstract/document/6795724}.

\bibitem[Letac \& Massam(2007)Letac and Massam]{letacWishartDistributionsDecomposable2007}
G{\'e}rard Letac and H{\'e}l{\`e}ne Massam.
\newblock Wishart distributions for decomposable graphs.
\newblock \emph{The Annals of Statistics}, 35\penalty0 (3):\penalty0 1278--1323, July 2007.
\newblock ISSN 0090-5364, 2168-8966.
\newblock \doi{10.1214/009053606000001235}.
\newblock URL \url{https://projecteuclid.org/journals/annals-of-statistics/volume-35/issue-3/Wishart-distributions-for-decomposable-graphs/10.1214/009053606000001235.full}.

\bibitem[Li et~al.(2024)Li, Tsai, Chen, and Liao]{liGraphNeuralNetworks2024}
Cheng-Te Li, Yu-Che Tsai, Chih-Yao Chen, and Jay~Chiehen Liao.
\newblock Graph {{Neural Networks}} for {{Tabular Data Learning}}: {{A Survey}} with {{Taxonomy}} and {{Directions}}, January 2024.
\newblock URL \url{http://arxiv.org/abs/2401.02143}.

\bibitem[Li et~al.(2019{\natexlab{a}})Li, Cui, Wu, Zhang, and Wang]{liFiGNNModelingFeature2020}
Zekun Li, Zeyu Cui, Shu Wu, Xiaoyu Zhang, and Liang Wang.
\newblock Fi-{{GNN}}: {{Modeling}} feature interactions via graph neural networks for {{CTR}} prediction.
\newblock In \emph{Proceedings of the 28th {{ACM}} International Conference on Information and Knowledge Management}, Cikm '19, pp.\  539--548, Beijing, China and New York, NY, USA, 2019{\natexlab{a}}. Association for Computing Machinery.
\newblock ISBN 978-1-4503-6976-3.
\newblock \doi{10.1145/3357384.3357951}.
\newblock URL \url{https://doi.org/10.1145/3357384.3357951}.

\bibitem[Li et~al.(2019{\natexlab{b}})Li, Cui, Wu, Zhang, and Wang]{liFiGNN_ArXiv_v1}
Zekun Li, Zeyu Cui, Shu Wu, Xiaoyu Zhang, and Liang Wang.
\newblock Fi-{{GNN}}: {{Modeling Feature Interactions}} via {{Graph Neural Networks}} for {{CTR Prediction}}, October 2019{\natexlab{b}}.
\newblock URL \url{https://arxiv.org/abs/1910.05552v1}.

\bibitem[Li et~al.(2020)Li, Cui, Wu, Zhang, and Wang]{liFiGNN_ArXiv_v2}
Zekun Li, Zeyu Cui, Shu Wu, Xiaoyu Zhang, and Liang Wang.
\newblock Fi-{{GNN}}: {{Modeling Feature Interactions}} via {{Graph Neural Networks}} for {{CTR Prediction}}, July 2020.
\newblock URL \url{http://arxiv.org/abs/1910.05552}.

\bibitem[Lian et~al.(2018)Lian, Zhou, Zhang, Chen, Xie, and Sun]{lianXDeepFMCombiningExplicit2018}
Jianxun Lian, Xiaohuan Zhou, Fuzheng Zhang, Zhongxia Chen, Xing Xie, and Guangzhong Sun.
\newblock {{xDeepFM}}: {{Combining Explicit}} and {{Implicit Feature Interactions}} for {{Recommender Systems}}.
\newblock In \emph{Proceedings of the 24th {{ACM SIGKDD International Conference}} on {{Knowledge Discovery}} \& {{Data Mining}}}, pp.\  1754--1763, London United Kingdom, July 2018. ACM.
\newblock ISBN 978-1-4503-5552-0.
\newblock \doi{10.1145/3219819.3220023}.
\newblock URL \url{https://dl.acm.org/doi/10.1145/3219819.3220023}.

\bibitem[Lindstrom \& Bates(1988)Lindstrom and Bates]{lindstromNewtonRaphsonEM1988}
Mary~J. Lindstrom and Douglas~M. Bates.
\newblock Newton---{{Raphson}} and {{EM Algorithms}} for {{Linear Mixed-Effects Models}} for {{Repeated-Measures Data}}.
\newblock \emph{Journal of the American Statistical Association}, 83\penalty0 (404):\penalty0 1014--1022, December 1988.
\newblock ISSN 0162-1459.
\newblock \doi{10.1080/01621459.1988.10478693}.
\newblock URL \url{https://doi.org/10.1080/01621459.1988.10478693}.

\bibitem[Lopardo et~al.(2024)Lopardo, Precioso, and Garreau]{lopardoAttentionMeetsPosthoc2024}
Gianluigi Lopardo, Frederic Precioso, and Damien Garreau.
\newblock Attention {{Meets Post-hoc Interpretability}}: {{A Mathematical Perspective}}, June 2024.
\newblock URL \url{http://arxiv.org/abs/2402.03485}.

\bibitem[Mann \& Whitney(1947)Mann and Whitney]{mannTestWhetherOne1947}
H.~B. Mann and D.~R. Whitney.
\newblock On a {{Test}} of {{Whether}} one of {{Two Random Variables}} is {{Stochastically Larger}} than the {{Other}}.
\newblock \emph{The Annals of Mathematical Statistics}, 18\penalty0 (1):\penalty0 50--60, 1947.
\newblock ISSN 0003-4851.
\newblock URL \url{https://www.jstor.org/stable/2236101}.

\bibitem[Marchetti et~al.(2024)Marchetti, Hillar, Kragic, and Sanborn]{marchettiHarmonicsLearningUniversal2024}
Giovanni~Luca Marchetti, Christopher~J. Hillar, Danica Kragic, and Sophia Sanborn.
\newblock Harmonics of {{Learning}}: {{Universal Fourier Features Emerge}} in {{Invariant Networks}}.
\newblock In \emph{Proceedings of {{Thirty Seventh Conference}} on {{Learning Theory}}}, pp.\  3775--3797. PMLR, June 2024.
\newblock URL \url{https://proceedings.mlr.press/v247/marchetti24a.html}.

\bibitem[McElfresh et~al.(2024)McElfresh, Khandagale, Valverde, C, Feuer, Hegde, Ramakrishnan, Goldblum, and White]{mcelfreshWhenNeuralNets2024}
Duncan McElfresh, Sujay Khandagale, Jonathan Valverde, Vishak~Prasad C, Benjamin Feuer, Chinmay Hegde, Ganesh Ramakrishnan, Micah Goldblum, and Colin White.
\newblock When {{Do Neural Nets Outperform Boosted Trees}} on {{Tabular Data}}?, July 2024.
\newblock URL \url{http://arxiv.org/abs/2305.02997}.

\bibitem[Mohammadi \& Wit(2015)Mohammadi and Wit]{mohammadiBayesianStructureLearning2015}
A.~Mohammadi and E.~C. Wit.
\newblock Bayesian {{Structure Learning}} in {{Sparse Gaussian Graphical Models}}.
\newblock \emph{Bayesian Analysis}, 10\penalty0 (1), March 2015.
\newblock ISSN 1936-0975.
\newblock \doi{10.1214/14-BA889}.
\newblock URL \url{http://arxiv.org/abs/1210.5371}.

\bibitem[Mohammadi \& Wit(2019)Mohammadi and Wit]{mohammadiBDgraphPackageBayesian2019}
Reza Mohammadi and Ernst~C. Wit.
\newblock {{BDgraph}}: {{An R Package}} for {{Bayesian Structure Learning}} in {{Graphical Models}}.
\newblock \emph{Journal of Statistical Software}, 89:\penalty0 1--30, May 2019.
\newblock ISSN 1548-7660.
\newblock \doi{10.18637/jss.v089.i03}.
\newblock URL \url{https://doi.org/10.18637/jss.v089.i03}.

\bibitem[Nadeau \& Bengio(1999)Nadeau and Bengio]{nadeauInferenceGeneralizationError1999}
Claude Nadeau and Yoshua Bengio.
\newblock Inference for the {{Generalization Error}}.
\newblock In \emph{Advances in {{Neural Information Processing Systems}}}, volume~12. MIT Press, 1999.
\newblock URL \url{https://papers.nips.cc/paper_files/paper/1999/hash/7d12b66d3df6af8d429c1a357d8b9e1a-Abstract.html}.

\bibitem[Padhi et~al.(2021)Padhi, Schiff, Melnyk, Rigotti, Mroueh, Dognin, Ross, Nair, and Altman]{padhiTabularTransformersModeling2021}
Inkit Padhi, Yair Schiff, Igor Melnyk, Mattia Rigotti, Youssef Mroueh, Pierre Dognin, Jerret Ross, Ravi Nair, and Erik Altman.
\newblock Tabular {{Transformers}} for {{Modeling Multivariate Time Series}}.
\newblock In \emph{{{ICASSP}} 2021 - 2021 {{IEEE International Conference}} on {{Acoustics}}, {{Speech}} and {{Signal Processing}} ({{ICASSP}})}, pp.\  3565--3569, June 2021.
\newblock \doi{10.1109/ICASSP39728.2021.9414142}.
\newblock URL \url{https://ieeexplore.ieee.org/abstract/document/9414142}.

\bibitem[Pearl(2021)]{pearlCausalityModelsReasoning2021}
Judea Pearl.
\newblock \emph{Causality: Models, Reasoning and Inference}.
\newblock Cambridge University Press, Cambridge, second edition, reprinted with corrections 2021 edition, 2021.
\newblock ISBN 978-0-511-80316-1.

\bibitem[Pinheiro \& Bates(2000)Pinheiro and Bates]{pinheiro2000mixed}
J.~Pinheiro and D.~Bates.
\newblock \emph{Mixed-Effects Models in {{S}} and s-{{PLUS}}}.
\newblock Statistics and Computing. Springer New York, 2000.
\newblock ISBN 978-0-387-98957-0.
\newblock URL \url{https://books.google.nl/books?id=N3WeyHFbHLQC}.

\bibitem[Prince(2023)]{prince2023understanding}
Simon~J.D. Prince.
\newblock \emph{Understanding Deep Learning}.
\newblock The MIT Press, 2023.
\newblock URL \url{http://udlbook.com}.

\bibitem[Prokhorenkova et~al.(2019)Prokhorenkova, Gusev, Vorobev, Dorogush, and Gulin]{prokhorenkovaCatBoostUnbiasedBoosting2019}
Liudmila Prokhorenkova, Gleb Gusev, Aleksandr Vorobev, Anna~Veronika Dorogush, and Andrey Gulin.
\newblock {{CatBoost}}: Unbiased boosting with categorical features, January 2019.
\newblock URL \url{http://arxiv.org/abs/1706.09516}.

\bibitem[Qu et~al.(2025)Qu, Holzm{\"u}ller, Varoquaux, and Morvan]{quTabICLTabularFoundation2025}
Jingang Qu, David Holzm{\"u}ller, Ga{\"e}l Varoquaux, and Marine~Le Morvan.
\newblock {{TabICL}}: {{A Tabular Foundation Model}} for {{In-Context Learning}} on {{Large Data}}, May 2025.
\newblock URL \url{http://arxiv.org/abs/2502.05564}.

\bibitem[Robinson et~al.(2024)Robinson, Ranjan, Hu, Huang, Han, Dobles, Fey, Lenssen, Yuan, Zhang, He, and Leskovec]{robinsonRelBenchBenchmarkDeep2024}
Joshua Robinson, Rishabh Ranjan, Weihua Hu, Kexin Huang, Jiaqi Han, Alejandro Dobles, Matthias Fey, Jan~E. Lenssen, Yiwen Yuan, Zecheng Zhang, Xinwei He, and Jure Leskovec.
\newblock {{RelBench}}: A benchmark for deep learning on relational databases.
\newblock In A.~Globerson, L.~Mackey, D.~Belgrave, A.~Fan, U.~Paquet, J.~Tomczak, and C.~Zhang (eds.), \emph{Advances in Neural Information Processing Systems}, volume~37, pp.\  21330--21341. Curran Associates, Inc., 2024.
\newblock URL \url{https://proceedings.neurips.cc/paper_files/paper/2024/file/25cd345233c65fac1fec0ce61d0f7836-Paper-Datasets_and_Benchmarks_Track.pdf}.

\bibitem[Romero~Guzman(2024)]{romeroguzmanGoodEfficientInductive2024}
David~Wilson Romero~Guzman.
\newblock \emph{The {{Good}}, the {{Efficient}} and the {{Inductive Biases}}:: {{Exploring Efficiency}} in {{Deep Learning Through}} the {{Use}} of {{Inductive Biases}}}.
\newblock {{PhD-Thesis}} - {{Research}} and graduation internal, Vrije Universiteit Amsterdam, September 2024.

\bibitem[Roverato(2002)]{roveratoHyperInverseWishart2002}
Alberto Roverato.
\newblock Hyper {{Inverse Wishart Distribution}} for {{Non-decomposable Graphs}} and its {{Application}} to {{Bayesian Inference}} for {{Gaussian Graphical Models}}.
\newblock \emph{Scandinavian Journal of Statistics}, 29\penalty0 (3):\penalty0 391--411, 2002.
\newblock ISSN 1467-9469.
\newblock \doi{10.1111/1467-9469.00297}.
\newblock URL \url{https://onlinelibrary.wiley.com/doi/abs/10.1111/1467-9469.00297}.

\bibitem[Somepalli et~al.(2021)Somepalli, Goldblum, Schwarzschild, Bruss, and Goldstein]{somepalliSAINTImprovedNeural2021}
Gowthami Somepalli, Micah Goldblum, Avi Schwarzschild, C.~Bayan Bruss, and Tom Goldstein.
\newblock {{SAINT}}: {{Improved Neural Networks}} for {{Tabular Data}} via {{Row Attention}} and {{Contrastive Pre-Training}}, June 2021.
\newblock URL \url{http://arxiv.org/abs/2106.01342}.

\bibitem[Song et~al.(2019)Song, Shi, Xiao, Duan, Xu, Zhang, and Tang]{songAutoIntAutomaticFeature2019}
Weiping Song, Chence Shi, Zhiping Xiao, Zhijian Duan, Yewen Xu, Ming Zhang, and Jian Tang.
\newblock {{AutoInt}}: {{Automatic Feature Interaction Learning}} via {{Self-Attentive Neural Networks}}, August 2019.
\newblock URL \url{http://arxiv.org/abs/1810.11921}.

\bibitem[Su et~al.(2024)Su, Ahmed, Lu, Pan, Bo, and Liu]{suRoFormerEnhancedTransformer2024}
Jianlin Su, Murtadha Ahmed, Yu~Lu, Shengfeng Pan, Wen Bo, and Yunfeng Liu.
\newblock {{RoFormer}}: {{Enhanced}} transformer with {{Rotary Position Embedding}}.
\newblock \emph{Neurocomputing}, 568:\penalty0 127063, February 2024.
\newblock ISSN 0925-2312.
\newblock \doi{10.1016/j.neucom.2023.127063}.
\newblock URL \url{https://www.sciencedirect.com/science/article/pii/S0925231223011864}.

\bibitem[Sundararajan et~al.(2017)Sundararajan, Taly, and Yan]{sundararajanAxiomaticAttributionDeep2017}
Mukund Sundararajan, Ankur Taly, and Qiqi Yan.
\newblock Axiomatic {{Attribution}} for {{Deep Networks}}, June 2017.
\newblock URL \url{http://arxiv.org/abs/1703.01365}.

\bibitem[Vaswani et~al.(2017)Vaswani, Shazeer, Parmar, Uszkoreit, Jones, Gomez, Kaiser, and Polosukhin]{vaswaniAttentionAllYou2017}
Ashish Vaswani, Noam Shazeer, Niki Parmar, Jakob Uszkoreit, Llion Jones, Aidan~N. Gomez, Lukasz Kaiser, and Illia Polosukhin.
\newblock Attention {{Is All You Need}}, December 2017.
\newblock URL \url{http://arxiv.org/abs/1706.03762}.

\bibitem[Vig(2019)]{vigMultiscaleVisualizationAttention2019}
Jesse Vig.
\newblock A {{Multiscale Visualization}} of {{Attention}} in the {{Transformer Model}}.
\newblock In Marta~R. {Costa-juss{\`a}} and Enrique Alfonseca (eds.), \emph{Proceedings of the 57th {{Annual Meeting}} of the {{Association}} for {{Computational Linguistics}}: {{System Demonstrations}}}, pp.\  37--42, Florence, Italy, July 2019. Association for Computational Linguistics.
\newblock \doi{10.18653/v1/P19-3007}.
\newblock URL \url{https://aclanthology.org/P19-3007/}.

\bibitem[{Villaiz{\'a}n-Vallelado} et~al.(2024){Villaiz{\'a}n-Vallelado}, Salvatori, Carro, and {Sanchez-Esguevillas}]{villaizan-valleladoGraphNeuralNetwork2024}
Mario {Villaiz{\'a}n-Vallelado}, Matteo Salvatori, Bel{\'e}n Carro, and Antonio~Javier {Sanchez-Esguevillas}.
\newblock Graph {{Neural Network}} contextual embedding for {{Deep Learning}} on tabular data.
\newblock \emph{Neural Networks}, 173:\penalty0 106180, May 2024.
\newblock ISSN 0893-6080.
\newblock \doi{10.1016/j.neunet.2024.106180}.
\newblock URL \url{https://www.sciencedirect.com/science/article/pii/S0893608024001047}.

\bibitem[Vogels et~al.(2024)Vogels, Mohammadi, Schoonhoven, and Birbil]{vogelsBayesianStructureLearning2024}
Lucas Vogels, Reza Mohammadi, Marit Schoonhoven, and {\c S}.~{\.I}lker Birbil.
\newblock Bayesian {{Structure Learning}} in {{Undirected Gaussian Graphical Models}}: {{Literature Review}} with {{Empirical Comparison}}.
\newblock \emph{Journal of the American Statistical Association}, 119\penalty0 (548):\penalty0 3164--3182, October 2024.
\newblock ISSN 0162-1459.
\newblock \doi{10.1080/01621459.2024.2395504}.
\newblock URL \url{https://doi.org/10.1080/01621459.2024.2395504}.

\bibitem[Wang et~al.(2017)Wang, Fu, Fu, and Wang]{wangDeepCrossNetwork2017}
Ruoxi Wang, Bin Fu, Gang Fu, and Mingliang Wang.
\newblock Deep \& {{Cross Network}} for {{Ad Click Predictions}}.
\newblock In \emph{Proceedings of the {{ADKDD}}'17}, pp.\  1--7, Halifax NS Canada, August 2017. ACM.
\newblock ISBN 978-1-4503-5194-2.
\newblock \doi{10.1145/3124749.3124754}.
\newblock URL \url{https://dl.acm.org/doi/10.1145/3124749.3124754}.

\bibitem[Wang et~al.(2020)Wang, Shivanna, Cheng, Jain, Lin, Hong, and Chi]{wangDCNV2Improved2020}
Ruoxi Wang, Rakesh Shivanna, Derek~Z. Cheng, Sagar Jain, Dong Lin, Lichan Hong, and Ed~H. Chi.
\newblock {{DCN V2}}: {{Improved Deep}} \& {{Cross Network}} and {{Practical Lessons}} for {{Web-scale Learning}} to {{Rank Systems}}, October 2020.
\newblock URL \url{http://arxiv.org/abs/2008.13535}.

\bibitem[Wilcoxon(1945)]{wilcoxonIndividualComparisonsRanking1945}
Frank Wilcoxon.
\newblock Individual {{Comparisons}} by {{Ranking Methods}}.
\newblock \emph{Biometrics Bulletin}, 1\penalty0 (6):\penalty0 80, December 1945.
\newblock ISSN 00994987.
\newblock \doi{10.2307/3001968}.
\newblock URL \url{https://www.jstor.org/stable/10.2307/3001968?origin=crossref}.

\bibitem[Wistuba et~al.(2015)Wistuba, Schilling, and {Schmidt-Thieme}]{wistubaLearningHyperparameterOptimization2015}
Martin Wistuba, Nicolas Schilling, and Lars {Schmidt-Thieme}.
\newblock Learning hyperparameter optimization initializations.
\newblock In \emph{2015 {{IEEE International Conference}} on {{Data Science}} and {{Advanced Analytics}} ({{DSAA}})}, pp.\  1--10, October 2015.
\newblock \doi{10.1109/DSAA.2015.7344817}.
\newblock URL \url{https://ieeexplore.ieee.org/abstract/document/7344817}.

\bibitem[Yan et~al.(2023)Yan, Chen, Wu, Chen, and Wu]{yanT2GFormerOrganizingTabular2023}
Jiahuan Yan, Jintai Chen, Yixuan Wu, Danny~Z. Chen, and Jian Wu.
\newblock {{T2G-Former}}: Organizing tabular features into relation graphs promotes heterogeneous feature interaction.
\newblock In \emph{Proceedings of the {{Thirty-Seventh AAAI Conference}} on {{Artificial Intelligence}} and {{Thirty-Fifth Conference}} on {{Innovative Applications}} of {{Artificial Intelligence}} and {{Thirteenth Symposium}} on {{Educational Advances}} in {{Artificial Intelligence}}}, volume~37 of \emph{{{AAAI}}'23/{{IAAI}}'23/{{EAAI}}'23}, pp.\  10720--10728. AAAI Press, February 2023.
\newblock ISBN 978-1-57735-880-0.
\newblock \doi{10.1609/aaai.v37i9.26272}.
\newblock URL \url{https://doi.org/10.1609/aaai.v37i9.26272}.

\bibitem[Ye et~al.(2024)Ye, Yu, Shen, Yu, and Zeng]{yeCrossFeatureInteractiveTabular2024}
Mang Ye, Yi~Yu, Ziqin Shen, Wei Yu, and Qingyan Zeng.
\newblock Cross-{{Feature Interactive Tabular Data Modeling With Multiplex Graph Neural Networks}}.
\newblock \emph{IEEE Transactions on Knowledge and Data Engineering}, pp.\  1--15, 2024.
\newblock ISSN 1558-2191.
\newblock \doi{10.1109/TKDE.2024.3440654}.
\newblock URL \url{https://ieeexplore.ieee.org/document/10631296/?arnumber=10631296}.

\bibitem[Zhang et~al.(2025)Zhang, Ren, Yu, Yuan, Wang, Li, Wu, Mo, Mao, Hao, Dai, Xu, Li, Zhang, He, Wang, Zhang, Xu, Li, Gao, Zou, Liu, Liu, Xu, Cheng, Li, Zhou, Li, Fan, Lin, Han, Li, Lu, Xue, Jiang, Wang, Wang, and Cui]{zhangLimiXUnleashingStructuredData2025}
Xingxuan Zhang, Gang Ren, Han Yu, Hao Yuan, Hui Wang, Jiansheng Li, Jiayun Wu, Lang Mo, Li~Mao, Mingchao Hao, Ningbo Dai, Renzhe Xu, Shuyang Li, Tianyang Zhang, Yue He, Yuanrui Wang, Yunjia Zhang, Zijing Xu, Dongzhe Li, Fang Gao, Hao Zou, Jiandong Liu, Jiashuo Liu, Jiawei Xu, Kaijie Cheng, Kehan Li, Linjun Zhou, Qing Li, Shaohua Fan, Xiaoyu Lin, Xinyan Han, Xuanyue Li, Yan Lu, Yuan Xue, Yuanyuan Jiang, Zimu Wang, Zhenlei Wang, and Peng Cui.
\newblock {{LimiX}}: {{Unleashing Structured-Data Modeling Capability}} for {{Generalist Intelligence}}, September 2025.
\newblock URL \url{http://arxiv.org/abs/2509.03505}.

\bibitem[Zheng et~al.(2023)Zheng, Peng, Dang, Zhu, Liu, Zhang, and Zhou]{zhengDeepTabularData2023}
Qinghua Zheng, Zhen Peng, Zhuohang Dang, Linchao Zhu, Ziqi Liu, Zhiqiang Zhang, and Jun Zhou.
\newblock Deep {{Tabular Data Modeling With Dual-Route Structure-Adaptive Graph Networks}}.
\newblock \emph{IEEE Transactions on Knowledge and Data Engineering}, 35\penalty0 (9):\penalty0 9715--9727, September 2023.
\newblock ISSN 1558-2191.
\newblock \doi{10.1109/TKDE.2023.3249186}.
\newblock URL \url{https://ieeexplore.ieee.org/document/10054100}.

\bibitem[Zhou et~al.(2021)Zhou, Cui, Hu, Zhang, Yang, Liu, Wang, Li, and Sun]{zhouGraphNeuralNetworks2021}
Jie Zhou, Ganqu Cui, Shengding Hu, Zhengyan Zhang, Cheng Yang, Zhiyuan Liu, Lifeng Wang, Changcheng Li, and Maosong Sun.
\newblock Graph {{Neural Networks}}: {{A Review}} of {{Methods}} and {{Applications}}, October 2021.
\newblock URL \url{http://arxiv.org/abs/1812.08434}.

\bibitem[Zhou et~al.(2022)Zhou, Liu, Chen, Li, Choi, and Hu]{zhouTable2GraphTransformingTabular2022}
Kaixiong Zhou, Zirui Liu, Rui Chen, Li~Li, Soo-Hyun Choi, and Xia Hu.
\newblock {{Table2Graph}}: {{Transforming Tabular Data}} to {{Unified Weighted Graph}}.
\newblock In \emph{Proceedings of the {{Thirty-First International Joint Conference}} on {{Artificial Intelligence}}}, pp.\  2420--2426, Vienna, Austria, July 2022. International Joint Conferences on Artificial Intelligence Organization.
\newblock ISBN 978-1-956792-00-3.
\newblock \doi{10.24963/ijcai.2022/336}.
\newblock URL \url{https://www.ijcai.org/proceedings/2022/336}.

\end{thebibliography}
\end{document}